\documentclass{article}

\usepackage[final]{nips_2017}

\usepackage[utf8]{inputenc} 
\usepackage[T1]{fontenc}    
\usepackage{hyperref}       
\usepackage{url}            
\usepackage{booktabs}       
\usepackage{amsfonts}       
\usepackage{nicefrac}       
\usepackage{microtype}      

\usepackage{times}
\usepackage{latexsym}
\usepackage[pdftex]{graphicx} 
\usepackage{color}
\usepackage{wrapfig}

\title{Describing Semantic Representations of Brain Activity Evoked by Visual Stimuli}

\author{
  Eri Matsuo\\ {\bf Ichiro Kobayashi}\\
  \small{Ochanomizu University}\\
  \small{2-1-1 Ohtsuka, Bunkyo-ku,}\\ 
  \small{Tokyo 112-8610, Japan.}\\
  {\tt  \small{g1220535@is.ocha.ac.jp}}\\
  {\tt  \small{koba@is.ocha.ac.jp}}
  \And
  Shinji Nishimoto\\ {\bf Satoshi Nishida}\\
  \small{National Institute of Information}\\
  \small{and\hspace{0.2mm}Communications\hspace{0.2mm}Technology}\\
  \small{1-4, Yamadaoka, Suita-shi,}\\
  \small{Osaka, 565-0871, Japan}\\
  {\tt  \small{nishimoto@nict.go.jp}}\\
  {\tt  \small{s-nishida@nict.go.jp}}\\
  \And
  Hideki Asoh\\
  \small{National Institute of Advanced}\\
  \small{Industrial Science and Technology}\\
  \small{2-3-26, Aomi, Koto-ku,}\\
  \small{Tokyo, 135-0064, Japan}\\
  {\tt  \small{h.asoh@aist.go.jp}}}

\begin{document}

\maketitle


\begin{abstract}


Quantitative modeling of human brain activity based on language representations has been actively studied in systems neuroscience. 
However, previous studies examined word-level representation, and little is known about whether we could recover structured sentences from brain activity. 
This study attempts to generate natural language descriptions of semantic contents from human brain activity evoked by visual stimuli. 
To effectively use a small amount of available brain activity data, our proposed method employs a pre-trained image-captioning network model using a deep learning framework. 
To apply brain activity to the image-captioning network, we train regression models that learn the relationship between brain activity and deep-layer image features. 
The results demonstrate that the proposed model can decode brain activity and generate descriptions using natural language sentences. 
We also conducted several experiments with data from different subsets of brain regions known to process visual stimuli. 
The results suggest that semantic information for sentence generations is widespread across the entire cortex.

\end{abstract}

\section{Introduction}
Quantitative analysis of semantic activities in the human brain is an area of active study.
With the development of machine learning methods and the application of such 
methods to natural language processing, many studies have attempted to 
interpret and represent brain activity with the semantics categories of words.
In this paper, we propose a deep learning method to describe 
semantic representations evoked by visual stimuli, i.e., higher order perception, with natural language sentences using functional magnetic resonance imaging (fMRI) brain data.
This requires a large amount of training data, i.e., brain activity data observed by fMRI.
However, assembling a large-scale brain activity dataset is difficult 
because observing brain activity data with fMRI is expensive and
each human brain is different in its size and shape.
To handle this problem, we propose a model that associates the image features of
the intermediate layer of a pre-trained caption-generation system 
with brain activity, which makes it possible to generate natural 
language descriptions of the semantic representation of brain activity.
We used three methods to train the corresponding relationships between 
image features and the brain activity data, i.e.,
ridge regression, three-layer neural networks, and five-layer deep neural
networks (DNN), and we compared the results.
We also conducted an experiment using brain activity data from 
specific brain regions that process visual stimuli rather than data 
from the whole cortex to reduce the dimensionality of the input data 
assuming that the dimensionality of brain activity data is too large to train the model.

\section{Related Work}
\subsection{Quantitative analysis of brain activity}
\vspace{-1mm}
Recently, many neuroscience studies have attempted to
quantitatively analyze the semantic representation of what a human
recalls using the fMRI data of brain activity evoked
by visual stimuli, such as natural movies and
images~\citep{mitchell2008,nishimoto2011,pereira2013,huth2012,stansbury2013,horikawa2013}.
\citet{stansbury2013} employed latent Dirichlet allocation \citep{blei2003} 
to assign semantic labels to still pictures using
natural language descriptions synchronized with the pictures and
discussed the relationships between the visual stimuli evoked by the
still pictures and brain activity. Based on these relationships, they 
constructed a model that classifies brain activity into semantic categories 
to reveal areas of the brain that deal with particular semantic categories.
\citet{cukur2013} estimated how people semantically change their recognition of objects from brain activity data in cases where the subject pays attention to objects in a video.
\citet{huth2012,huth2016_dialog,huth2016_movie} revealed 
the corresponding relationships between brain activities and visual
stimuli using the semantic categories of WordNet~\citep{Miller1995}. 
They used the categories to construct a map for semantic representation 
in the cerebral cortex and showed that semantic information is represented in 
various patterns over broad areas of the cortex.
\citet{nishida2015} showed that, compared to other language models such as Latent Semantic Indexing \citep{deerwester1990} and latent Dirichlet allocation, the word2vec (skip-gram) model by \citet{mikolov2013} gives better accuracy in modeling the semantic representation of human brain activity.
Furthermore, they showed that there is a correlation between the 
distributed semantics obtained based on skip-gram in the word2vec 
framework with the Japanese Wikipedia corpus and brain activity observed 
through blood oxygen level dependent (BOLD) contrast obtained using fMRI.
As these studies indicate, statistical language models used to analyze 
semantic representation in human brain activity can explain higher order
cognitive representations.
In this context, although most previous studies use the semantic categories 
of words to describe the semantic representation of brain activity, 
in this study, we aim to take a further step toward quantitative
analysis of this relationship using a caption-generation system.
With this approach, we can describe brain activity with natural 
language sentences, which can represent richer information than words.
In addition, \citet{guclu} demonstrated that DNN trained for
action recognition can be used to predict how the dorsal stream, a brain
area that processes visual stimuli, responds to natural movies.
They demonstrated a correlation between the intermediate representations 
of DNN layers and dorsal stream areas.
This suggests the effectiveness of deep learning methods for the analysis 
and modeling of brain activities evoked by visual stimuli.

\vspace{-2mm}

\subsection{Caption generation from images}
\vspace{-1mm}
In natural language processing, deep learning has resulted in significant progress in natural language generation and machine translation. In particular, encoder-decoder (enc-dec) models, e.g., sequence-to-sequence models, have been studied~\citep{rnn_lm,encdec_lm1,encdec_lm2,encdec2,att_trans,att_abst}.
For example, such models have been applied to speech recognition~\citep{att_speech}, video captioning~\citep{att_video}, and text summarization~\citep{encdec_textsum}.
Typically, two approaches have been used for image-caption generation. The first approach retrieves and ranks existing captions~\citep{reuse1,reuse2,reuse3,reuse4}, and the second fills sentence templates based on features extracted from a given image~\citep{tmpl1,tmpl2,tmpl3,tmpl4}. 
Recently, methods that employ an enc-dec framework to generate captions for images have been proposed~\citep{encdec1,encdec3,encdec4,show_tell}.
\citet{show_tell} developed an image-caption generation method  by
building an enc-dec network employing GoogLeNet~\citep{lenet}, which
can extract image features effectively, as an encoder and a 
Long Short-Term Memory Language Model (LSTM-LM)~\citep{lstm,rnn_lm},
which is a deep neural language model, as an decoder.
In this study, we build and train an enc-dec network based on 
those prior studies~\cite{show_tell,show_attend,imcap}. However, we use brain activity data as input rather than images and attempt to generate natural language descriptions for the brain data.

\section{Proposed Method}
\begin{figure*}[t]
\begin{center}
  \includegraphics[width=14cm]{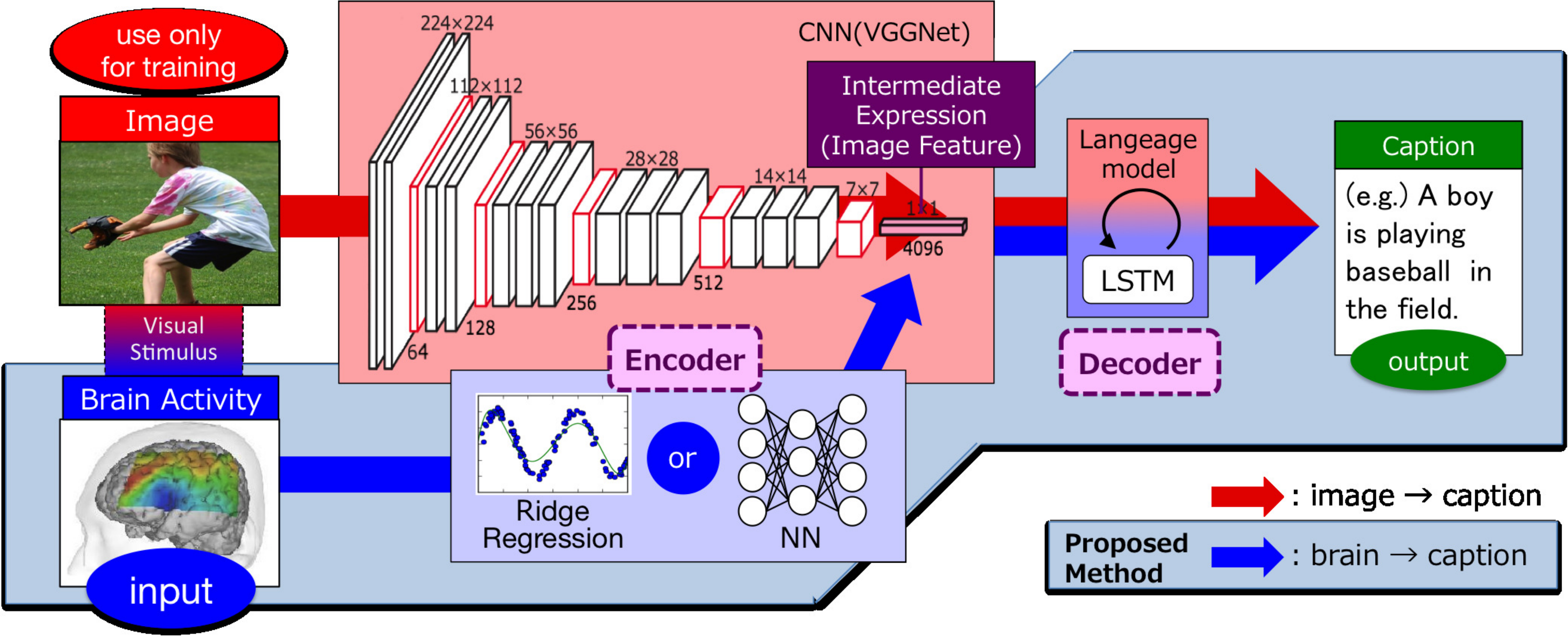}
  \vspace{-8mm}
  \caption{Overview of the proposed method.}
  \label{fig:abstract}
\end{center}
\vspace{-5mm}
\end{figure*}

This study aims to generate natural language sentences that describe
what a human being calls to mind using brain activity data observed
by fMRI as input information. We combined a image $\rightarrow$ caption model and a brain activity data $\rightarrow$ image feature model (Sections \ref{subsec:imcap} and \ref{subsec:brimf}).
Figure \ref{fig:abstract} presents an overview of the proposed method.

\subsection{Image $\rightarrow$ caption model (A)}  \label{subsec:imcap}

We employed an image-captioning model (A) based on a DNN framework, i.e., the enc-dec network~\citep{att_abst,show_tell}, as the main component of the proposed model.
In the enc-dec framework, by combining two DNN models
functioning as an encoder and a decoder, the model encodes input
information as an intermediate expression and decodes the information 
as an expression in a different modality.
Generally, previous studies of image-captioning systems have proposed 
enc-dec models that combine two DNNs: one DNN extracts 
image features using a convolutional neural network and the
other generates captions using a LSTM with the image features which correspond to an intermediate expression of the model.
Similar to such previous models, we constructed an image $\rightarrow$ image feature $\rightarrow$ caption model (A) employing VGGNet~\citep{vggnet} as an encoder and a two-layer LSTM language model~\citep{lstm,rnn_lm} as a decoder. 
We used pairs of image and caption data as training data.

\subsection{Brain activity data $\rightarrow$ image feature model (B)}  \label{subsec:brimf}

To apply the above image-captioning process to handle brain activity
data rather than images, we constructed a model that predicts features
extracted by VGGNet from images that evoke visual stimuli in the brain using fMRI
brain activity data as input. In other words, the model encodes brain activity data into the intermediate expression in the image $\rightarrow$ caption model (A).
We implemented and compared three models, i.e., a ridge regression model, a three-layer neural network model, and a five-layer DNN model, to determine which machine learning method is suitable for this model.
The five-layer DNN model was pre-trained using stacking
autoencoders~\citep{stackedAE} to avoid delay and overfitting in training
due to the lack of brain data.
We used pairs of fMRI brain activity data and the images a subject
observed as training data; however, we did not use natural language descriptions.

\section{Process Flow}
The process of the proposed method is as follows.
\begin{description}
{\small
\item{\bf Step 1.}\ {\em Encode brain activity as an intermediate expression.}\\
Model (B) predicts the feature of the image a subject watches from the input brain activity data evoked by the visual stimuli.
In the followings step, the features are provided to model (A) and processed as the intermediate expression. 
\item{\bf Step 2-1.}\ {\em Word estimation by the LSTM-LM.}\\
The LSTM-LM decoder of model (A) predicts the next word from the image feature produced in Step 1 and the hidden states of LSTM at the previous time step.
\item{\bf Step 2-2.}\ {\em Caption generation by iterative word estimation.}\\
A caption is generated by sequentially estimating words by repeating Step 2-1 until either the length of the sentence exceeds a predefined maximum or the terminal symbol of a sentence is output.
}
\end{description}
As mentioned above, we construct a brain activity data $\rightarrow$ image feature $\rightarrow$ caption model (C) by training the brain activity data $\rightarrow$ image feature model (B) and the image $\rightarrow$ image feature $\rightarrow$ caption model (A) individually and execute them sequentially in the prediction phase. 
Note that model (C) uses only fMRI brain activity data as input, i.e., without images. 

\section{Experiments}
Chainer\footnote{http://chainer.org/} was used as the deep learning framework to construct the neural networks.

\subsection{Experiment (A): image $\rightarrow$ caption model}
\subsubsection{Experimental settings}
Microsoft COCO\footnote{http://mscoco.org/}, which includes 414,113 pairs of images and their captions, was used as the dataset for the experiments.
The hyper-parameters of the models used in the experiments were set based on previous studies~\cite{show_tell,image_qa,encdec_lm1}.
The parameters to be learned were initialized by random values
obtained based on standard normal distribution.
The initial parameters of the word embedding layer were those used by word2vec learned using a skip-gram with window size = 5.
The pre-trained synaptic weights and hyper-parameters were used for VGGNet.
Furthermore, 3,469 words that appear more than 50 times in the
training data were used to generate natural language descriptions.
The settings for learning are shown in the leftmost column of 
Table \ref{tab:settings}.

\begin{table*}[t]
  \begin{center}
    \vspace{-6mm}
    \caption{Experimental settings}
   \label{tab:settings}
    \vspace{-2mm}
       \scalebox{0.64}{
       \begin{tabular}{c||c||c|c|c}
	& image $\rightarrow$ image feature $\rightarrow$ caption model & \multicolumn{3}{c}{brain activity data $\rightarrow$ image feature model} \\
	& & 1: Ridge Regression & 2: three-layer NN & 3: five-layer DNN \\ \hline\hline
	Dataset & Microsoft COCO & \multicolumn{3}{c}{brain activity data} \\ \hline
	Training quantity & 414,113 samples$\times$100 epochs & \multicolumn{3}{c}{4,500 samples$\times$1,000 epochs} \\ \hline
	Algorithm & Adam & Ridge regression & \multicolumn{2}{c}{stochastic gradient descent} \\ \hline
	 & a = 0.001, b1 = 0.9, b2 = 0.999, eps = 1e-8 & & \multicolumn{2}{c}{learning rate : 0.01} \\
	Hyper-parameters & gradient clipping threshold : 1  & L2-norm : 0.5 & \multicolumn{2}{c}{gradient clipping threshold : 1} \\
	 & L2-norm : 0.005 & & \multicolumn{2}{c}{L2-norm : 0.005} \\ \hline
	 & word embedding: word2vec & \multicolumn{2}{c|}{} & pre-trained with Stacked Autoencoder \\
	Initial parameters & VGGNet: pre-trained \&  & \multicolumn{2}{c|}{std normal random numbers} & using unsupervised brain activity data \\
	 & the others: std normal random numbers & \multicolumn{2}{c|}{} & ( 7,540 samples$\times$200 epochs )  \\ \hline
	Units per layer & 512 & 65,665-4,096 & 65,665-8,000-4,096 & 65,665-7,500-6,500-5,500-4,096\\ \hline
	Vocabulary & Frequent 3,469 words & \multicolumn{3}{c}{-} \\ \hline
	Loss function & cross entropy & \multicolumn{3}{c}{mean squared error} \\ \hline
       \end{tabular}
      }
   \vspace{-4mm}
 \end{center}
\end{table*}


\begin{figure}[b]
  \begin{center}
    \vspace{-3mm}
       \scalebox{0.83}{
       \begin{tabular}{clclc|c}
	\begin{minipage}{10.2mm}
	  \scalebox{0.14}{\includegraphics{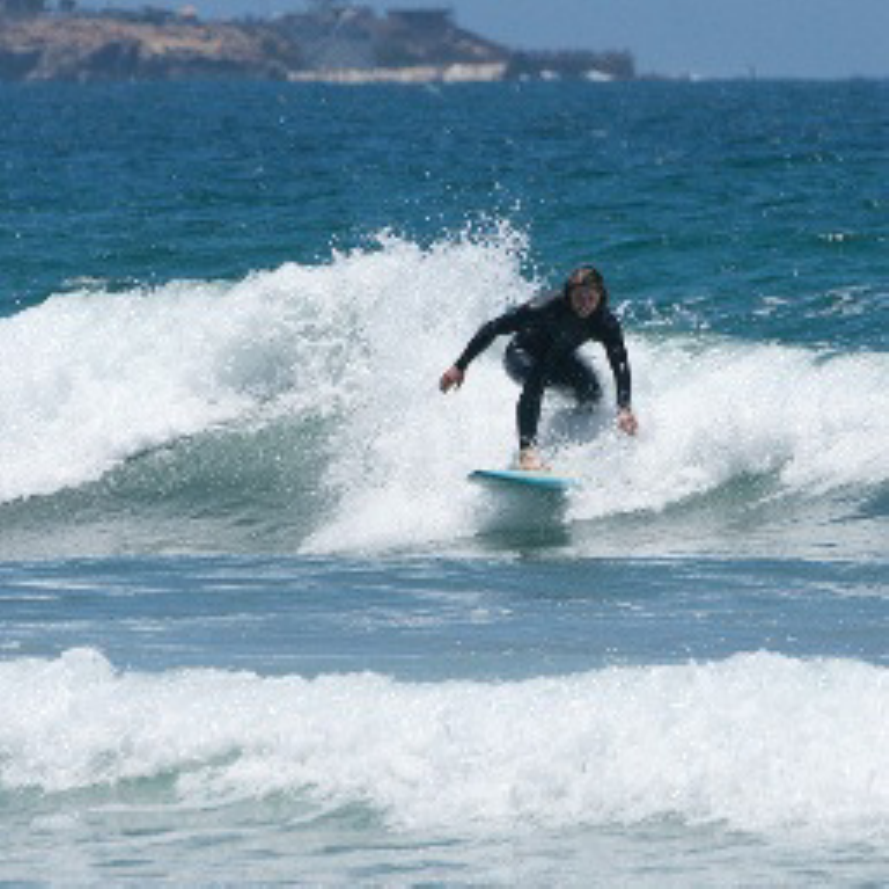}}
	\end{minipage} & A man is surfing in the ocean on his surfboard. &
	\begin{minipage}{10.2mm}
	  \scalebox{0.14}{\includegraphics{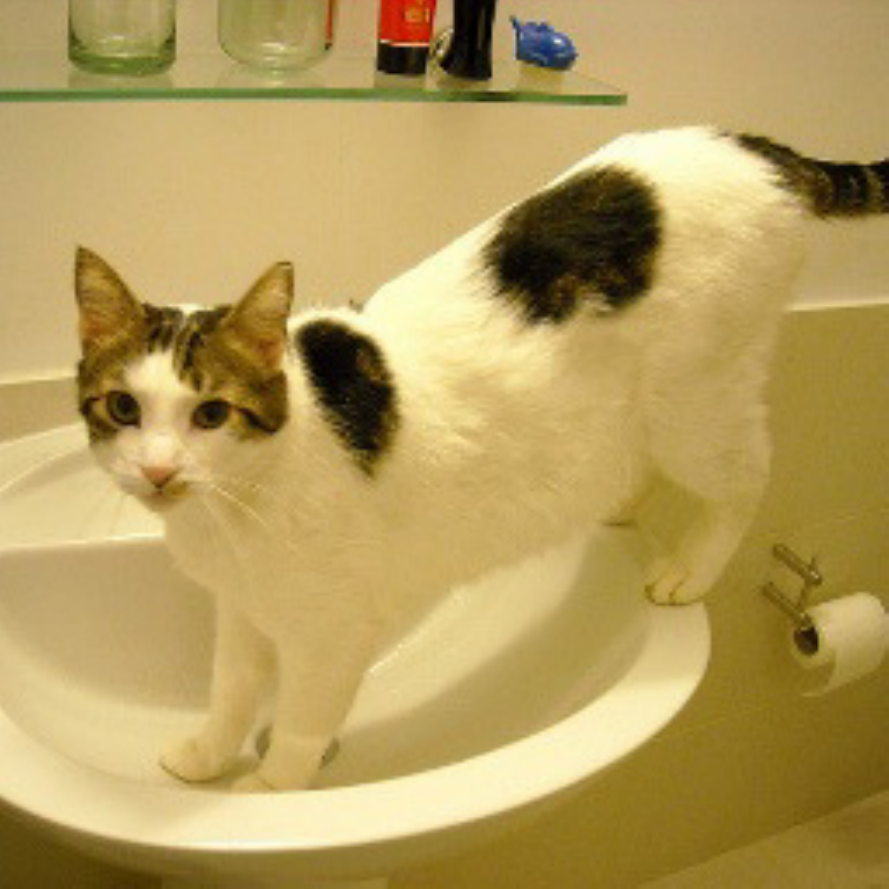}}
	\end{minipage} & A black and white cat is sitting on the toilet. 
       \end{tabular}
      }
    \vspace{-1mm}
    \caption{Captions for randomly selected images\label{fig:exp_a_imcap}}
 \vspace{-2mm}
 \end{center}
\end{figure}

\subsubsection{Results \& Discussion}


We confirmed the process of learning by the convergence of the perplexity of
output sentences recorded for each epoch. Figure \ref{fig:exp_a_imcap} 
shows the natural language descriptions for two images randomly selected from test images.
In the first example, a considerably reasonable natural language description was
generated. In the second example,
appropriate expressions for the subject of the generated
sentence, i.e., a cat, and its color were selected.
Furthermore, the prepositions, e.g., in and on, and articles, e.g., 
a and an, were also used correctly. Therefore, the output
sentences correctly describe the content of the images. 
As shown above, reasonable captions were generated for the 
test images and the perplexity converged near 2.5; therefore, 
an appropriate model was built to generate natural language descriptions from the images.
Note that errors that frequently appear in the generated sentences seem to depend on image processing rather than language processing. For instance, in the second example, standing is expressed as sitting and a washbasin is expressed as a toilet.

\begin{figure}[t]
  \begin{center}
    \vspace{-6mm}
       \scalebox{0.71}{
       \begin{tabular}{c c p{12.5em} p{12.5em} p{12.5em}}
	& Stimuli & \multicolumn{1}{c}{Ridge Regression} & \multicolumn{1}{c}{Three-layer NN} & \multicolumn{1}{c}{Five-layer DNN} \\ \hline
	\begin{minipage}{0.1mm}
	  \centering \scalebox{0.23}{\includegraphics{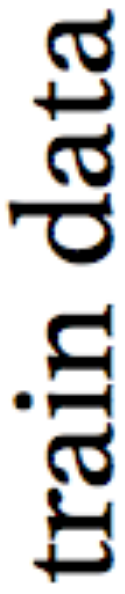}}
	\end{minipage}
	& \begin{minipage}{13mm}
	  \centering \scalebox{0.18}{\includegraphics{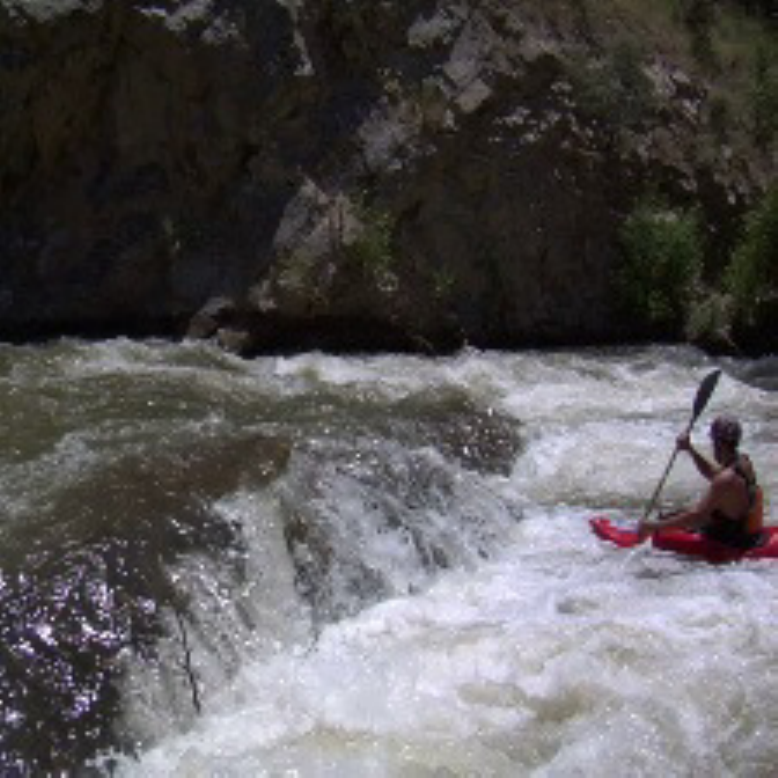}}
	\end{minipage}
	& \begin{minipage}{45mm}
	  \centering \scalebox{0.15}{\includegraphics{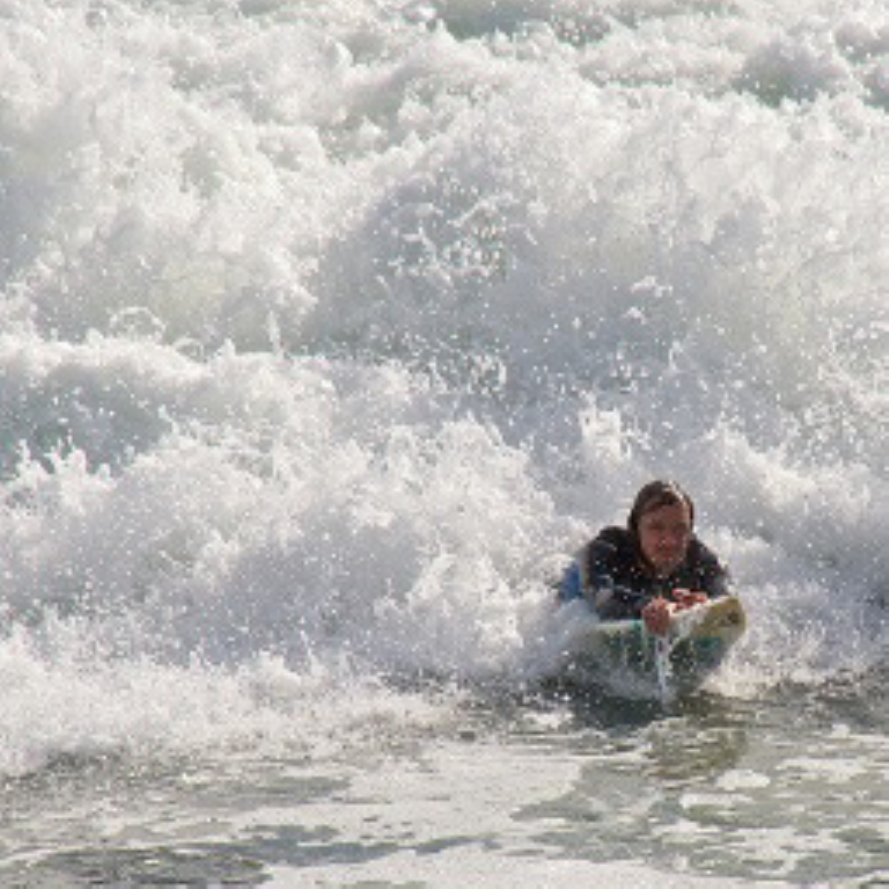}} 
	   \scalebox{0.15}{\includegraphics{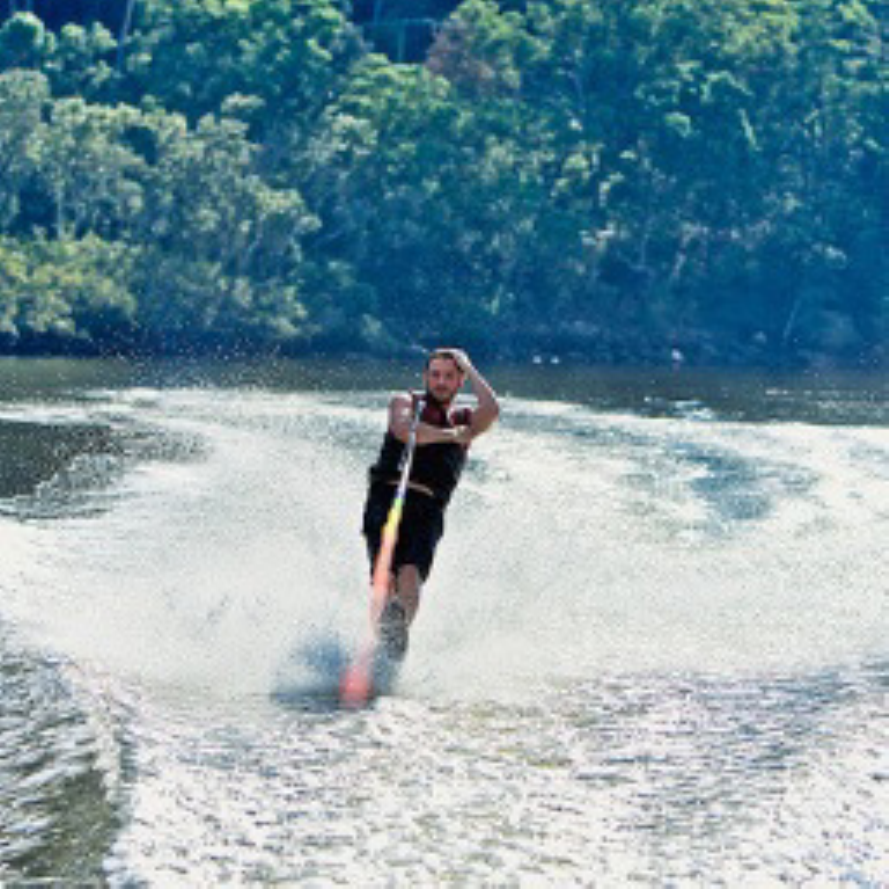}} 
	   \scalebox{0.15}{\includegraphics{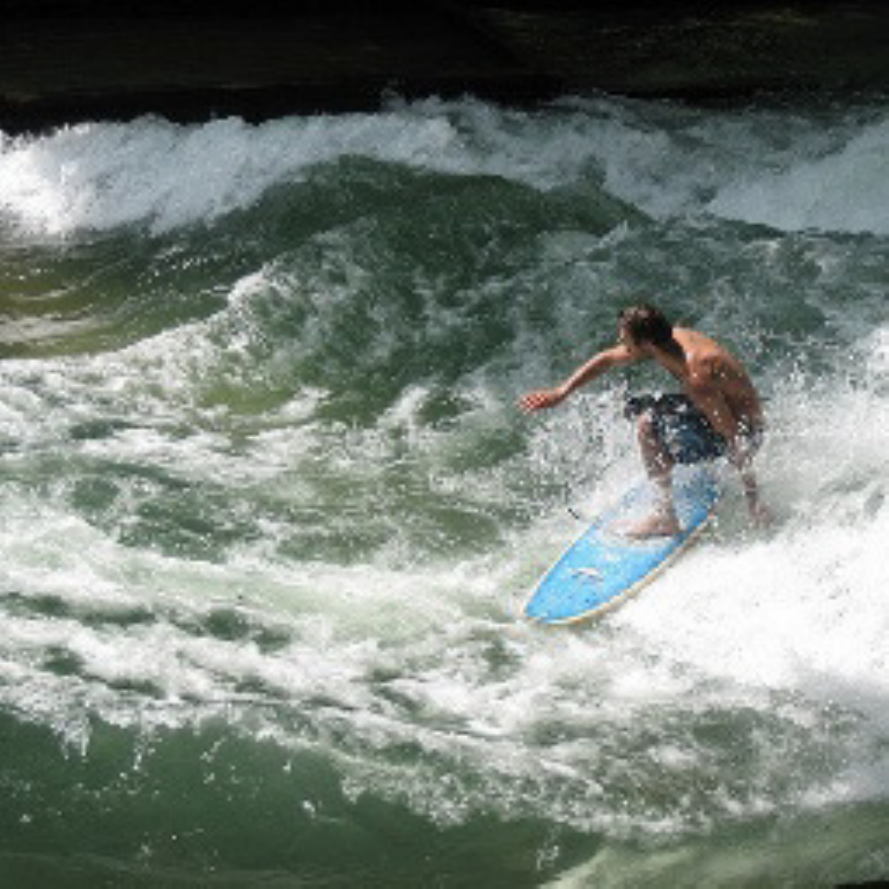}} 
	\end{minipage}
	& \begin{minipage}{45mm}
	  \centering \scalebox{0.15}{\includegraphics{figures/retrieve/train323_1.pdf}} 
	   \scalebox{0.15}{\includegraphics{figures/retrieve/train323_2.pdf}} 
	   \scalebox{0.15}{\includegraphics{figures/retrieve/train323_3.pdf}} 
	\end{minipage}
	& \begin{minipage}{45mm}
	  \centering \scalebox{0.15}{\includegraphics{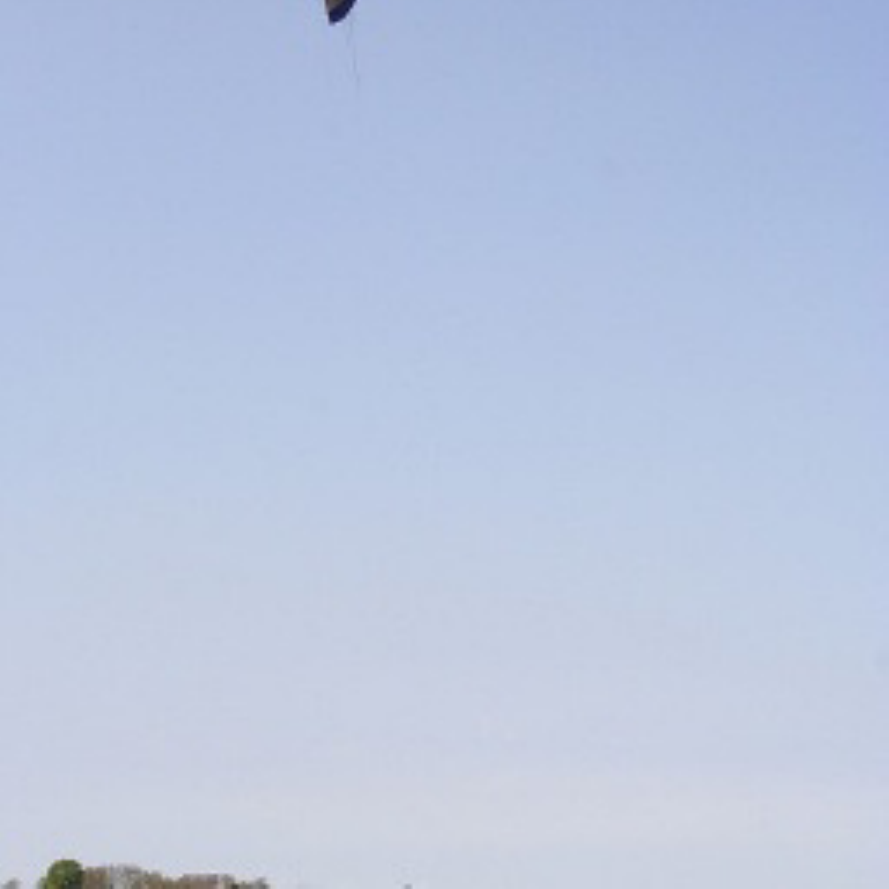}} 
	   \scalebox{0.15}{\includegraphics{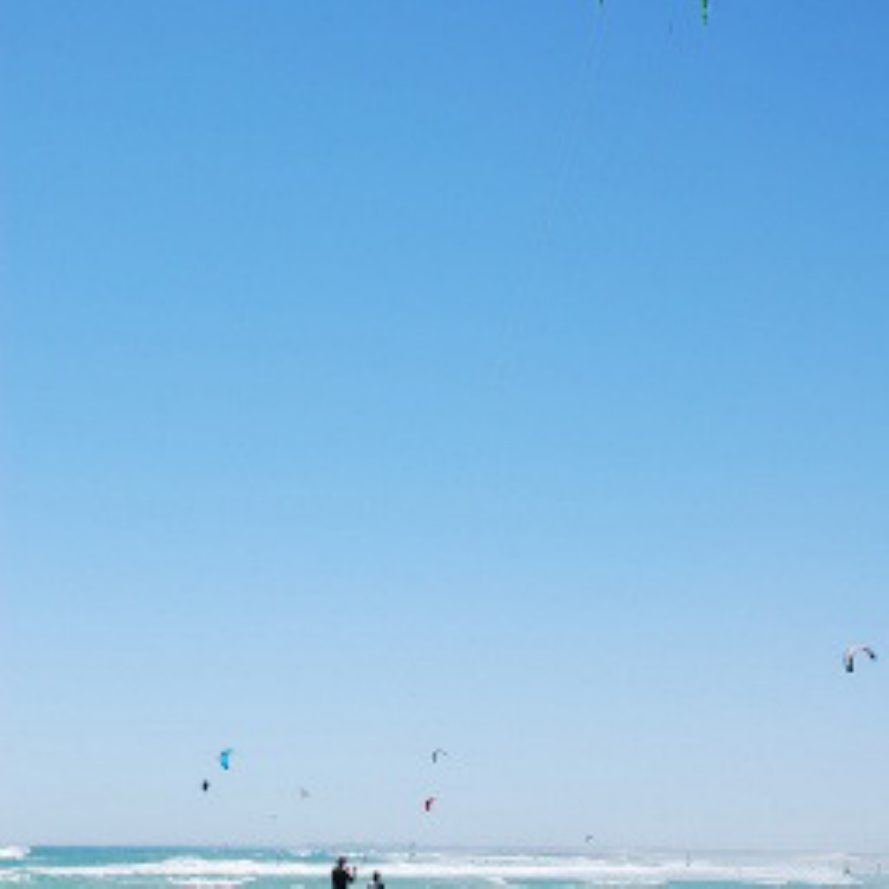}} 
	   \scalebox{0.15}{\includegraphics{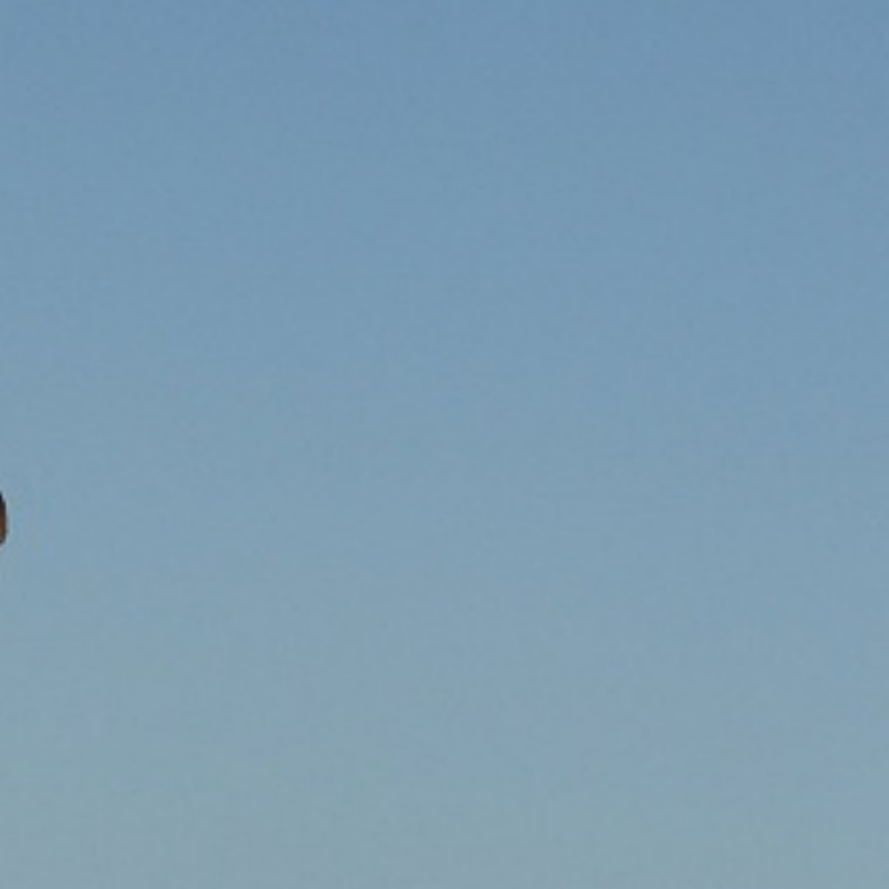}} 
	\end{minipage}\\
	& \begin{minipage}{13mm}
	  \centering \scalebox{0.18}{\includegraphics{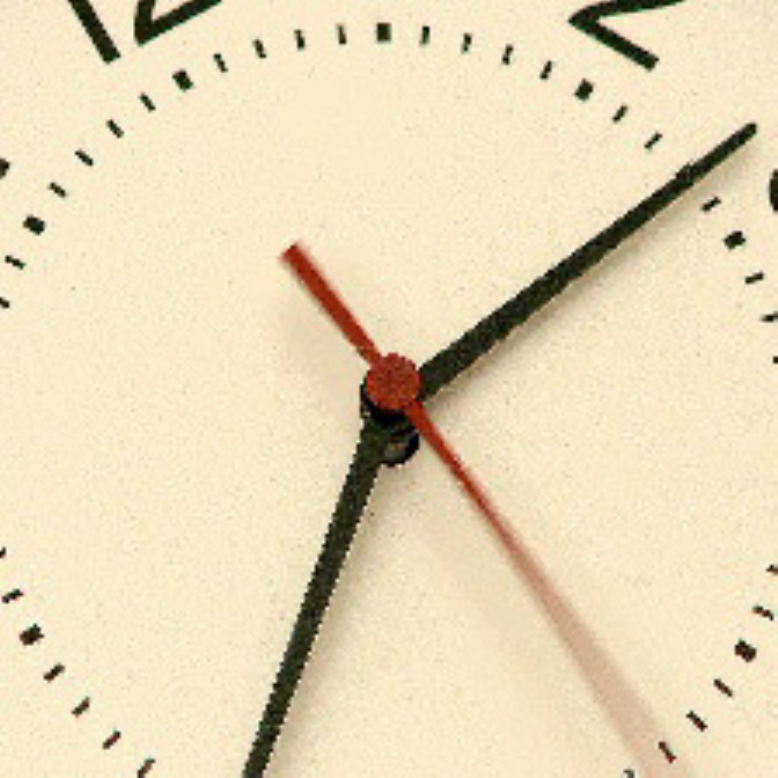}}
	\end{minipage}
	& \begin{minipage}{45mm}
	  \centering \scalebox{0.15}{\includegraphics{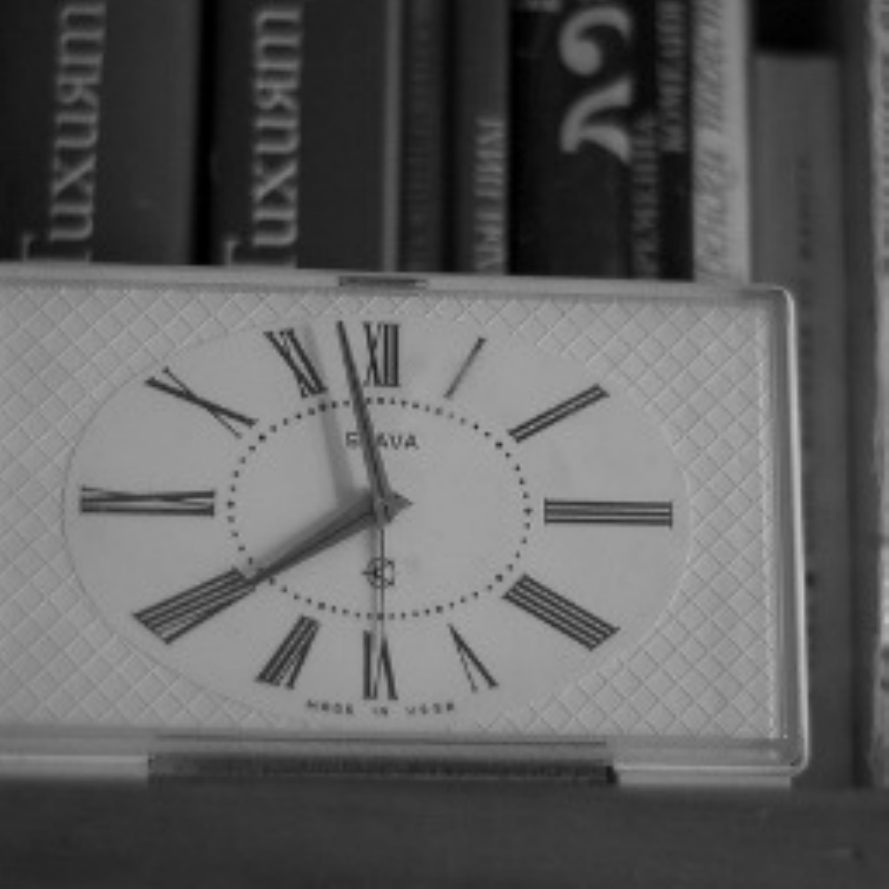}} 
	   \scalebox{0.15}{\includegraphics{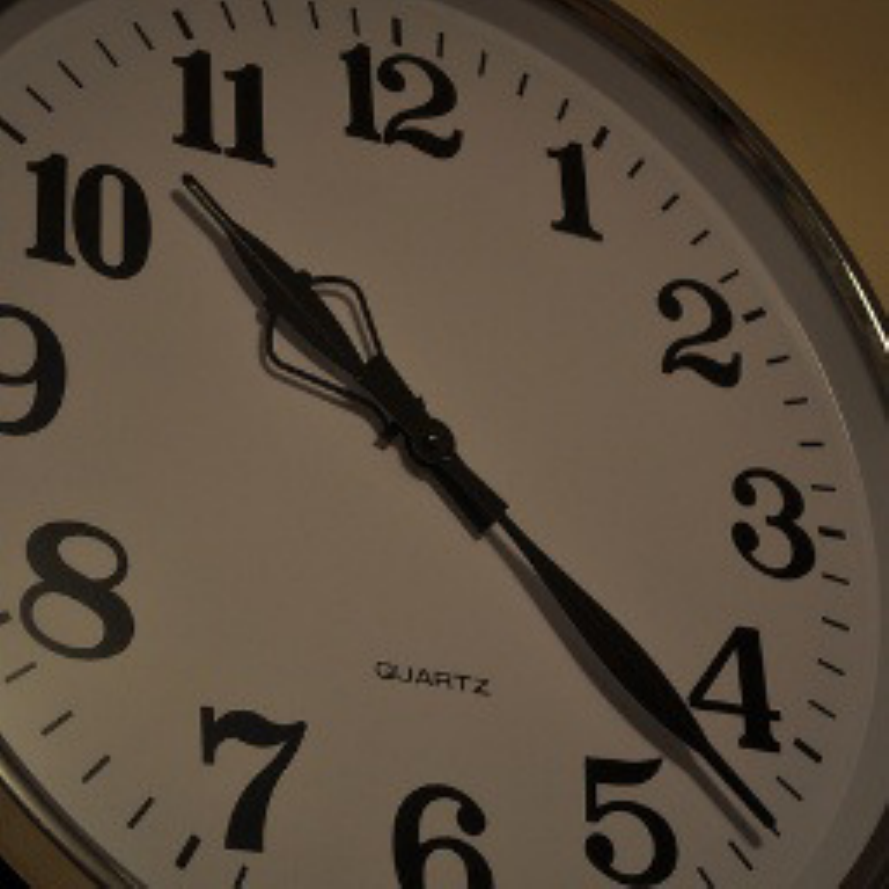}} 
	   \scalebox{0.15}{\includegraphics{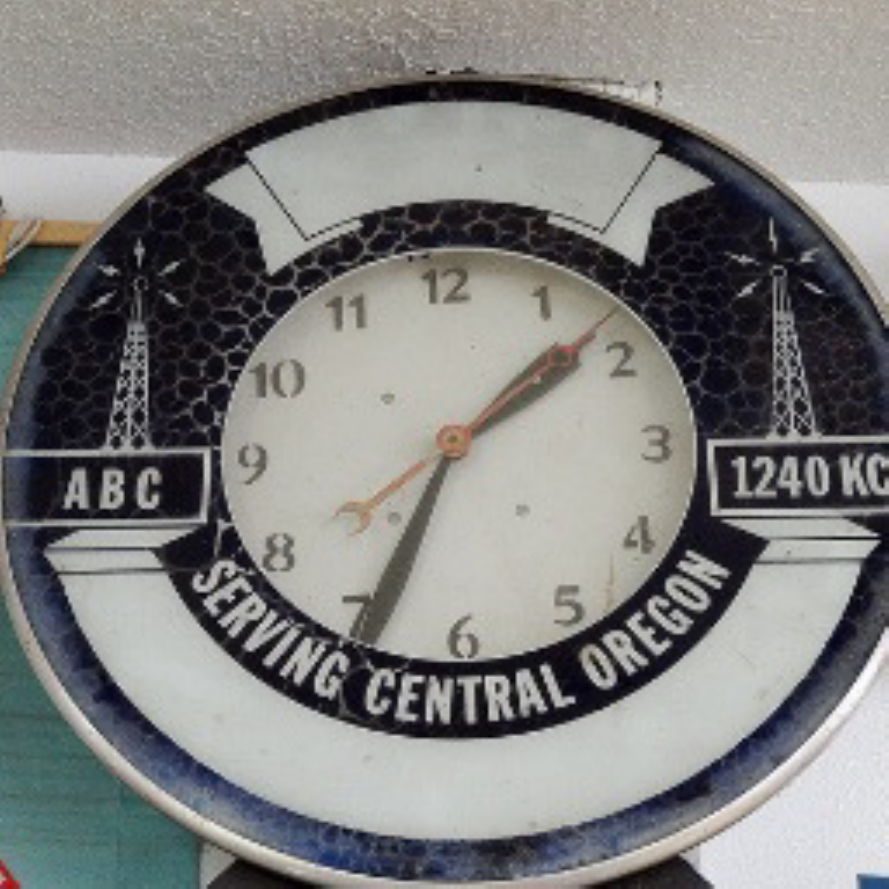}} 
	\end{minipage}
	& \begin{minipage}{45mm}
	  \centering \scalebox{0.15}{\includegraphics{figures/retrieve/train100_1.pdf}} 
	   \scalebox{0.15}{\includegraphics{figures/retrieve/train100_2.pdf}} 
	   \scalebox{0.15}{\includegraphics{figures/retrieve/train100_3.pdf}} 
	\end{minipage}
	& \begin{minipage}{45mm}
	  \centering \scalebox{0.15}{\includegraphics{figures/retrieve/dnn5_1.pdf}} 
	   \scalebox{0.15}{\includegraphics{figures/retrieve/dnn5_2.pdf}} 
	   \scalebox{0.15}{\includegraphics{figures/retrieve/dnn5_3.pdf}} 
	\end{minipage}\\ \hline
	\begin{minipage}{0.1mm}
	  \centering \scalebox{0.23}{\includegraphics{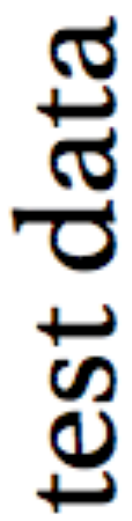}}
	\end{minipage}
	& \begin{minipage}{13mm}
	  \centering \scalebox{0.18}{\includegraphics{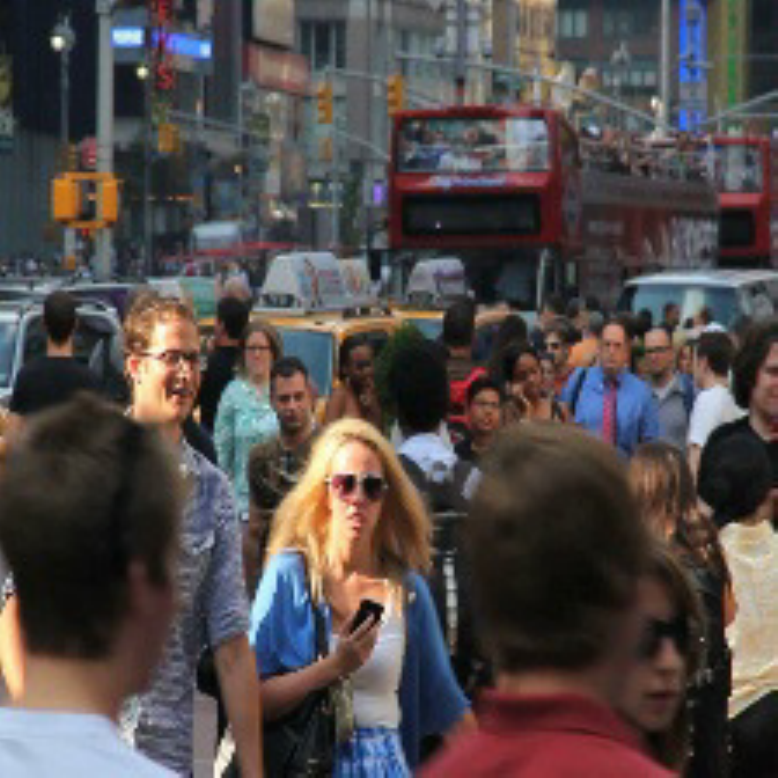}}
	\end{minipage} 
	& \begin{minipage}{45mm}
	  \centering \scalebox{0.15}{\includegraphics{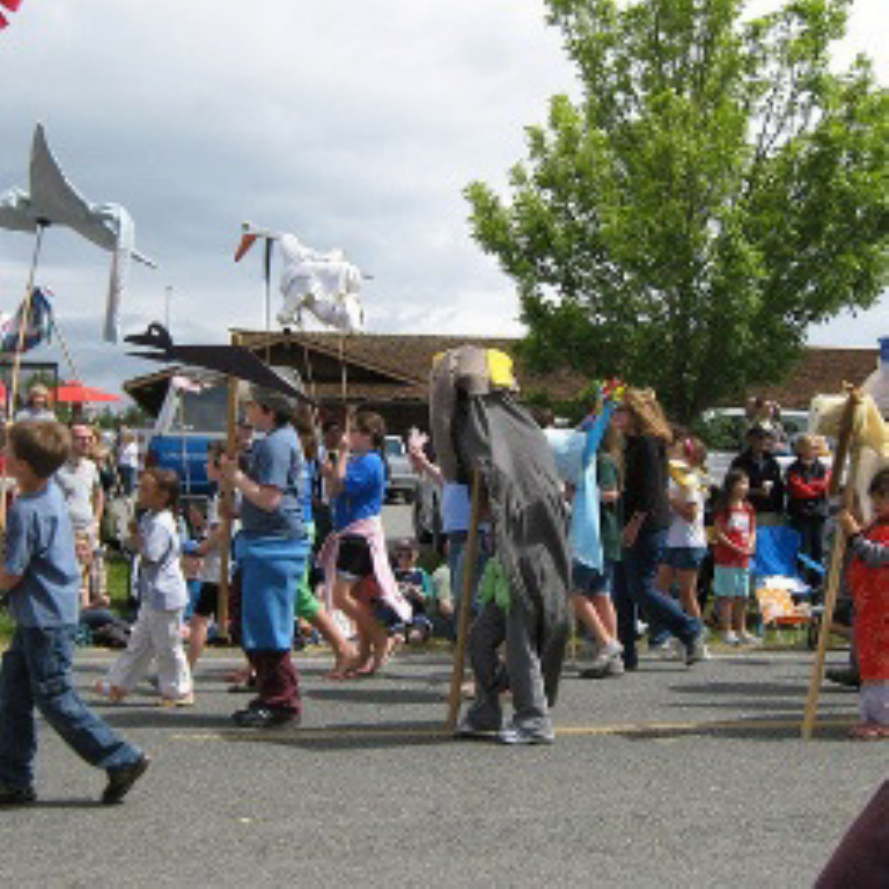}} 
	   \scalebox{0.15}{\includegraphics{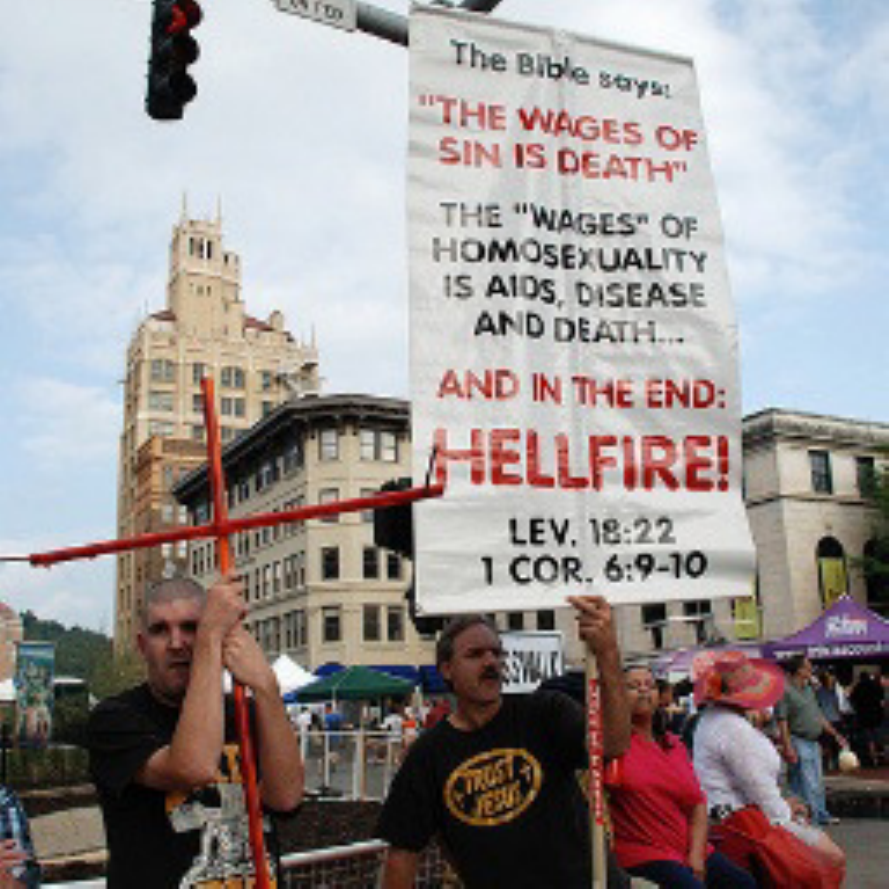}} 
	   \scalebox{0.15}{\includegraphics{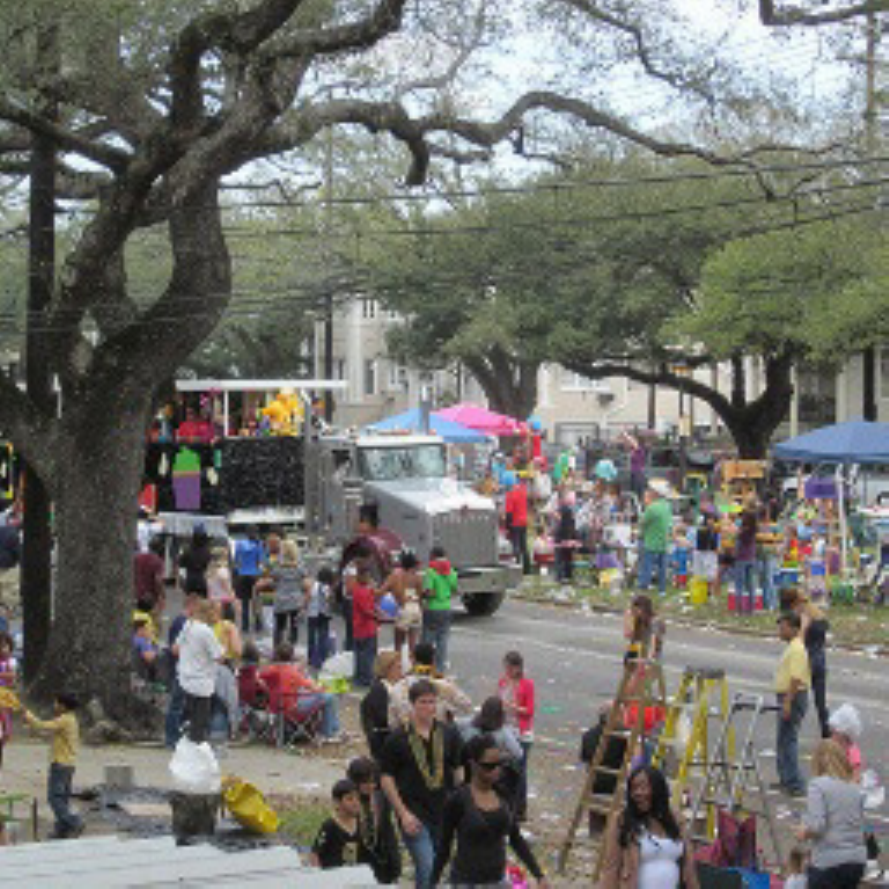}} 
	\end{minipage}
	& \begin{minipage}{45mm}
	  \centering \scalebox{0.15}{\includegraphics{figures/retrieve/test216_1.pdf}} 
	   \scalebox{0.15}{\includegraphics{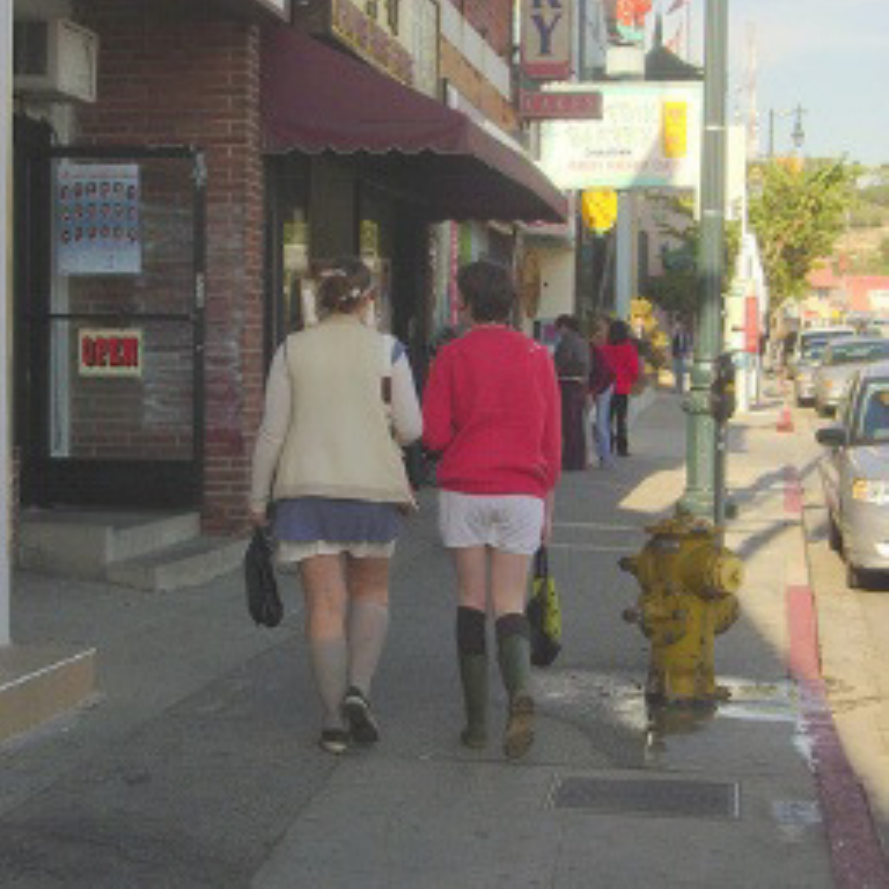}} 
	   \scalebox{0.15}{\includegraphics{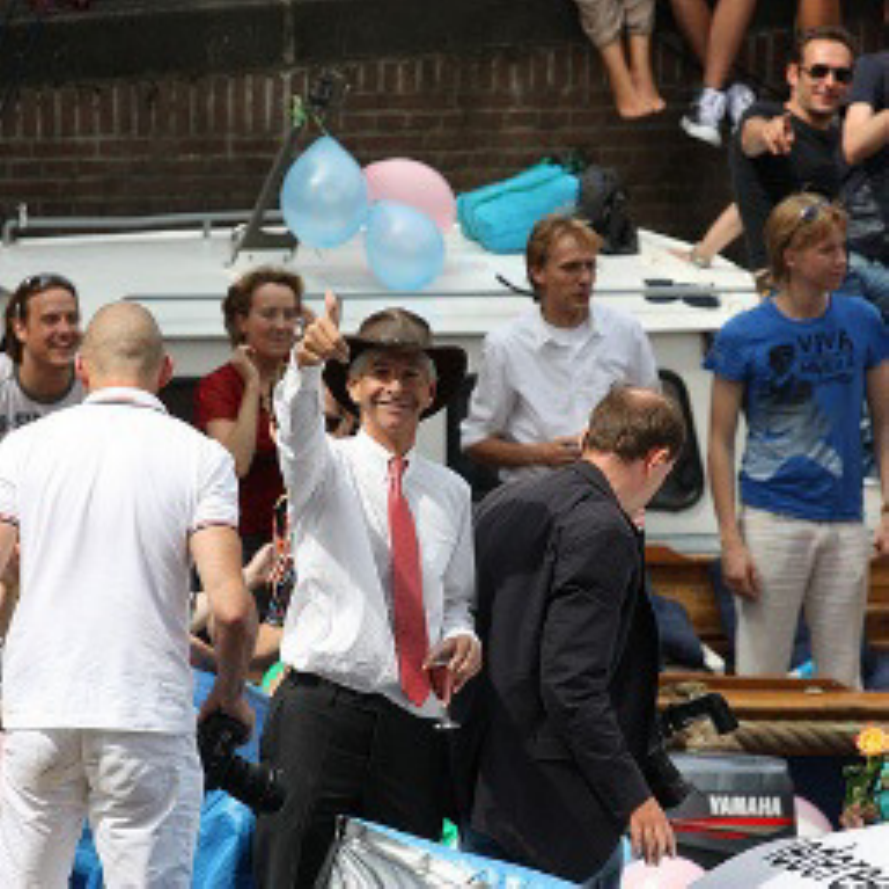}} 
	\end{minipage}
	& \begin{minipage}{45mm}
	  \centering \scalebox{0.15}{\includegraphics{figures/retrieve/dnn5_1.pdf}} 
	   \scalebox{0.15}{\includegraphics{figures/retrieve/dnn5_2.pdf}} 
	   \scalebox{0.15}{\includegraphics{figures/retrieve/dnn5_3.pdf}} 
	\end{minipage}\\
	& \begin{minipage}{13mm}
	  \centering \scalebox{0.18}{\includegraphics{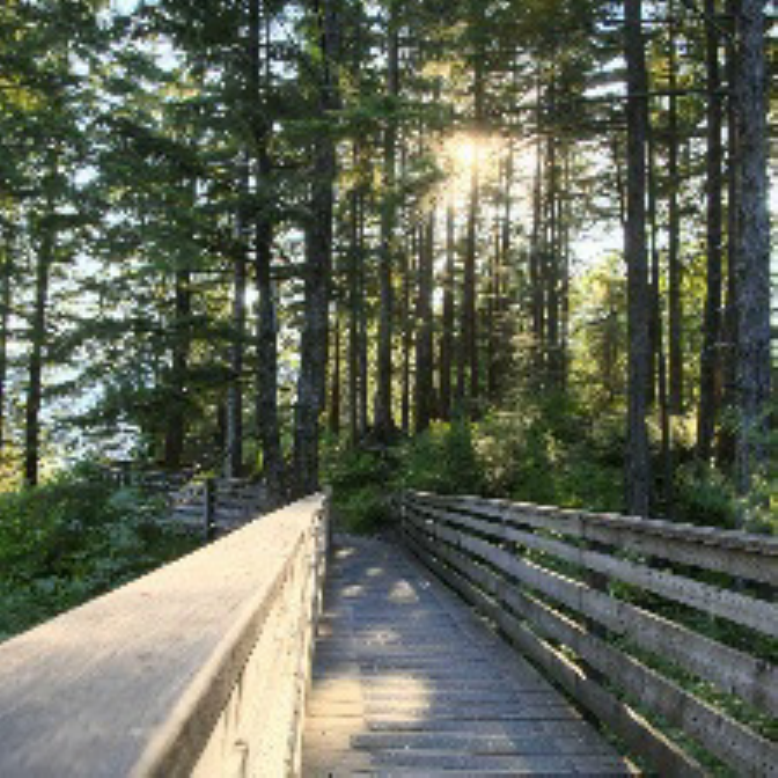}}
	\end{minipage} 
	& \begin{minipage}{45mm}
	  \centering \scalebox{0.15}{\includegraphics{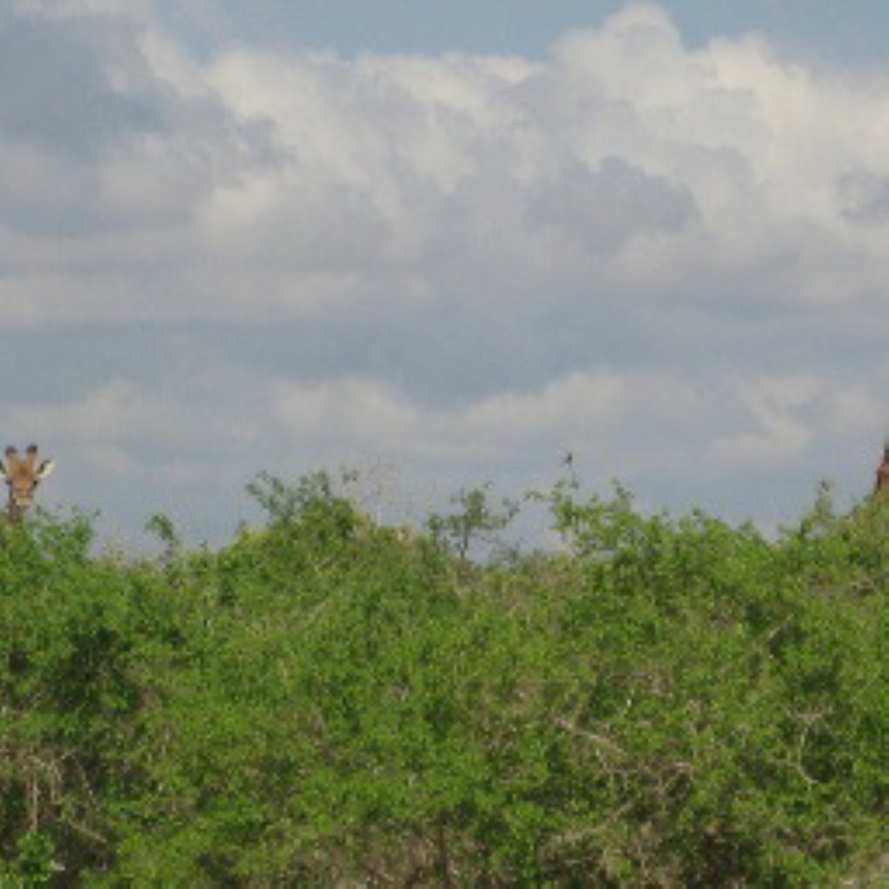}} 
	   \scalebox{0.15}{\includegraphics{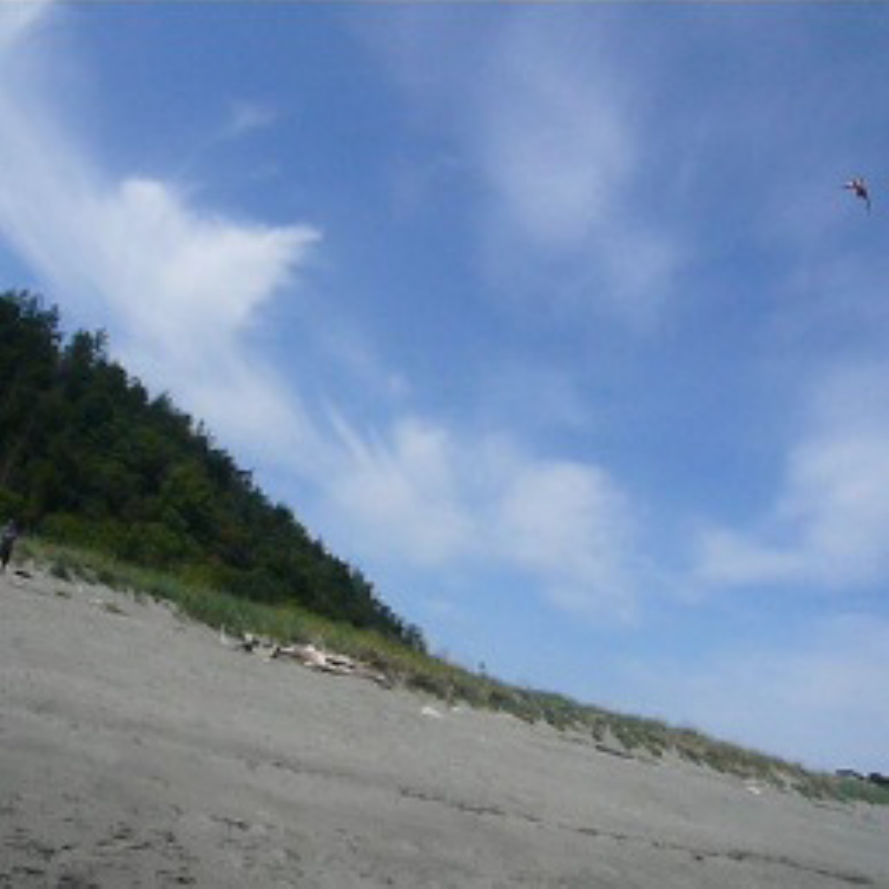}} 
	   \scalebox{0.15}{\includegraphics{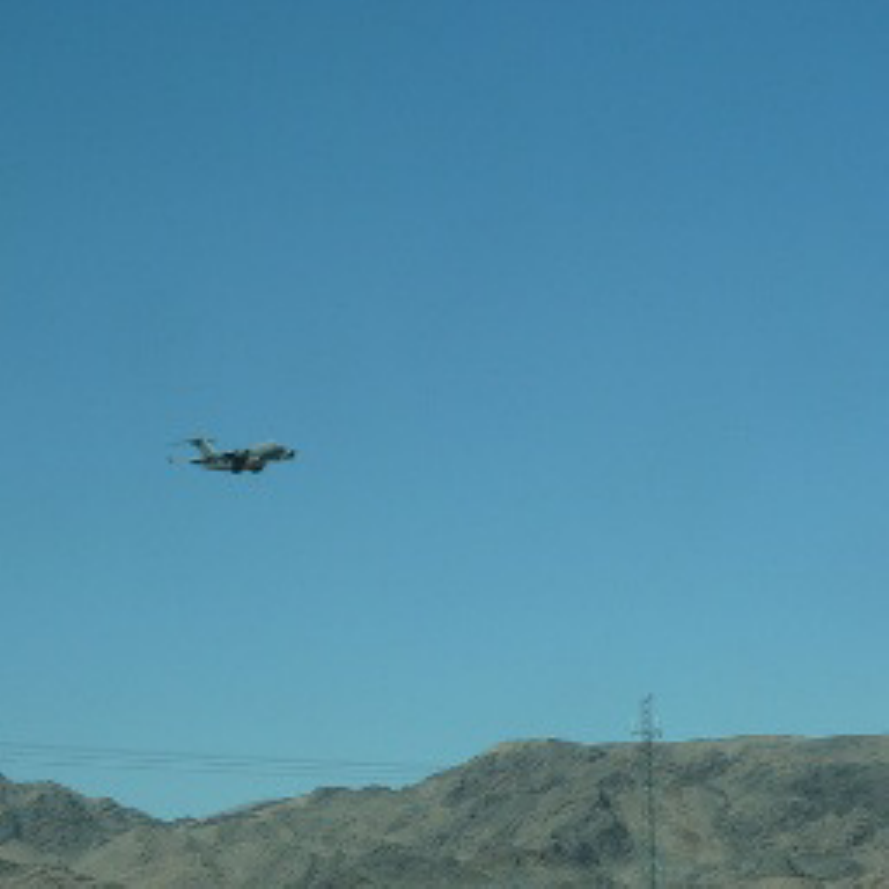}} 
	\end{minipage}
	& \begin{minipage}{45mm}
	  \centering \scalebox{0.15}{\includegraphics{figures/retrieve/test43_1.pdf}} 
	   \scalebox{0.15}{\includegraphics{figures/retrieve/test43_2.pdf}} 
	   \scalebox{0.15}{\includegraphics{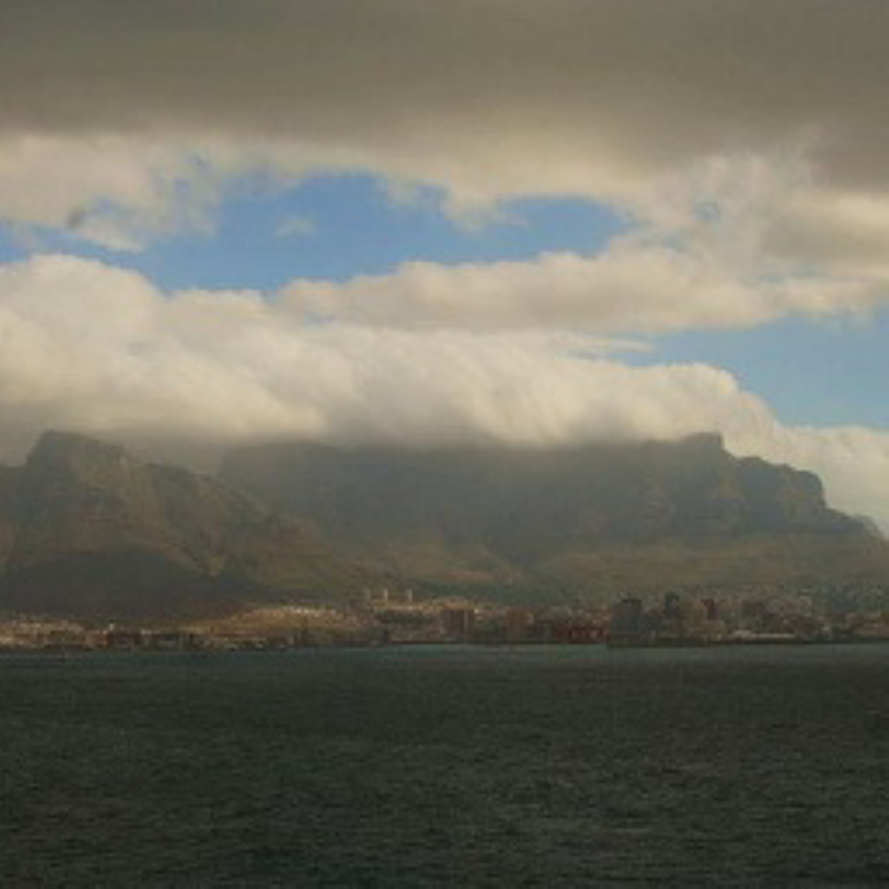}} 
	\end{minipage}
	& \begin{minipage}{45mm}
	  \centering \scalebox{0.15}{\includegraphics{figures/retrieve/dnn5_1.pdf}} 
	   \scalebox{0.15}{\includegraphics{figures/retrieve/dnn5_2.pdf}} 
	   \scalebox{0.15}{\includegraphics{figures/retrieve/dnn5_3.pdf}} 
	\end{minipage}\\
       \end{tabular}
      }
    \caption{Exp. (B-1): Stimulation image and retrieved similar images (top-3)\label{fig:exp_b1_retrieve}}
 \vspace{-5mm}
 \end{center}
\end{figure}

\subsection{Experiment (B-1): brain $\rightarrow$ image feature model}
\subsubsection{Experimental settings}
\begin{figure*}[h]
 \begin{center}  
  \includegraphics[width=10cm]{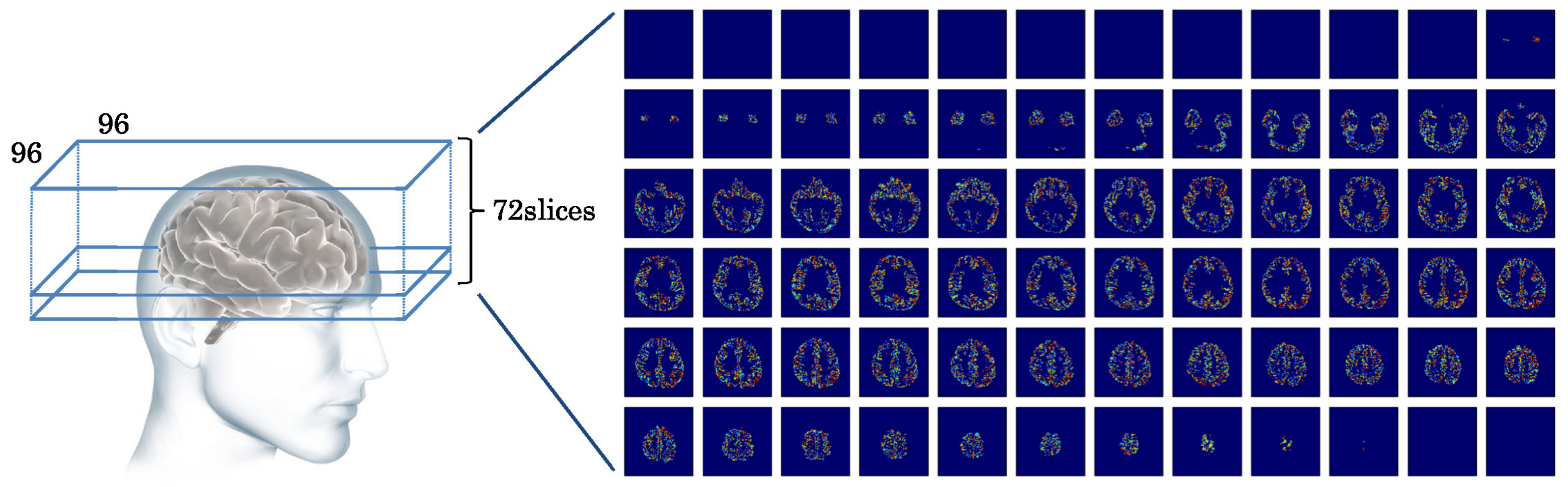}
  \vspace{-3mm}
  \caption{65,665 voxels observed by fMRI as the cerebral cortex region.}
  \label{fig:brain_data}
  \vspace{-4mm}
 \end{center}
\end{figure*}

For the learning dataset for the corresponding relationships between brain
activity and image features, we employed 
the brain activity data of a subject stimulated by natural movies
\citep{nishimoto2011}, i.e., the BOLD signal observed by fMRI, 
and still pictures taken from the movies provided as visual stimuli, 
which were synchronized with the brain activity data.
In the natural movies, there are various kinds of movies about natural phenomenon, 
artifacts, humans, films, 3D animations, etc., whose length of time are a few tens of seconds. 
As the input data, we employed 65,665 voxels corresponding to
the cerebral cortex part among 96$\times$96$\times$72
voxels observed by fMRI, (see, Figure \ref{fig:brain_data}), 
then learned the corresponding relationships between the brain data and the 
image features, whose dimensionality is 4,096, extracted from the image using VGGNet.
We used 4,500 samples as training data (recorded every two seconds for
9,000 seconds), which is a small number for learning a DNN.
In terms of ridge regression and the three-layer neural network model, 
the parameters to be learned were initialized with random values
obtained from the normal standard distribution. For the five-layer
DNN model, to increase the speed of learning and avoid overfitting,
we used 7,540 unlabeled brain data samples to pre-train the networks using
autoencoders for 200 epochs per layer, and we used the obtained
synaptic weights as the initial network parameters. The learning settings are 
shown in the three columns on the right side of Table \ref{tab:settings}. 

\subsubsection{Results \& Discussion}


We recorded the mean squared error (MSE) for each epoch and confirmed that the MSEs of the three models converged. 
For evaluation, we conducted an experiment to retrieve the images, which have similar
image features to those estimated from brain activity data, from
82,783 images of the Microsoft COCO training dataset with MSE metric. 
Figure \ref{fig:exp_b1_retrieve} shows the result of retrieving similar images.
As for ridge regression and three-layer neural network, those models retrieved
proper images from most training data, so we confirmed that the models 
could extract proper image features from brain activity
data. Furthermore, it also could be said that the images similar to the
stimulation images which evoked brain activities 
were retrieved from the image database.
However, as for five-layer DNN, the same unrelated images were
retrieved for all input brain activity data, and the results for the
test samples were worse than those for training samples even when employing ridge regression or the three-layer neural network although the MSEs for the test dataset converged.
The reason for this is that the input dimension, i.e., 65,665, was
much larger than that of the parameters to be learned, and the number of
training data samples, i.e., 4,500, was small. 
As a result, overfitting occurred due to the lack of adjustment of hyper-parameters.

\subsection{Experiment (C-1): brain $\rightarrow$ caption model}
\subsubsection{Experimental settings}
We built a model that generates a natural language description from brain activity data by combining 
the model in Experiment (A) and the three models in Experiment (B-1). We then
generated descriptions based on the three methods, i.e., ridge
regression, the three-layer neural network, and the five-layer DNN. In addition, we generated captions from the same images used in the above experiments with model (A).

\begin{figure}[t]
  \begin{center}
    \vspace{-6mm}
   \label{fig:exp_c1_brcap}
       \scalebox{0.71}{
       \begin{tabular}{c c p{11em} p{11em} p{12em} | p{11em}}
	& Stimuli & \multicolumn{1}{c}{Ridge Regression} & \multicolumn{1}{c}{Three-layer NN} & \multicolumn{1}{c}{Five-layer DNN} & \multicolumn{1}{c}{Image $\rightarrow$ Caption Model} \\ \hline
	\begin{minipage}{0.1mm}
	  \centering \scalebox{0.23}{\includegraphics{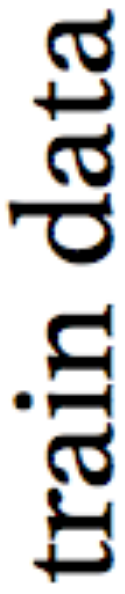}}
	\end{minipage}
	& \begin{minipage}{13mm}
	  \centering \scalebox{0.18}{\includegraphics{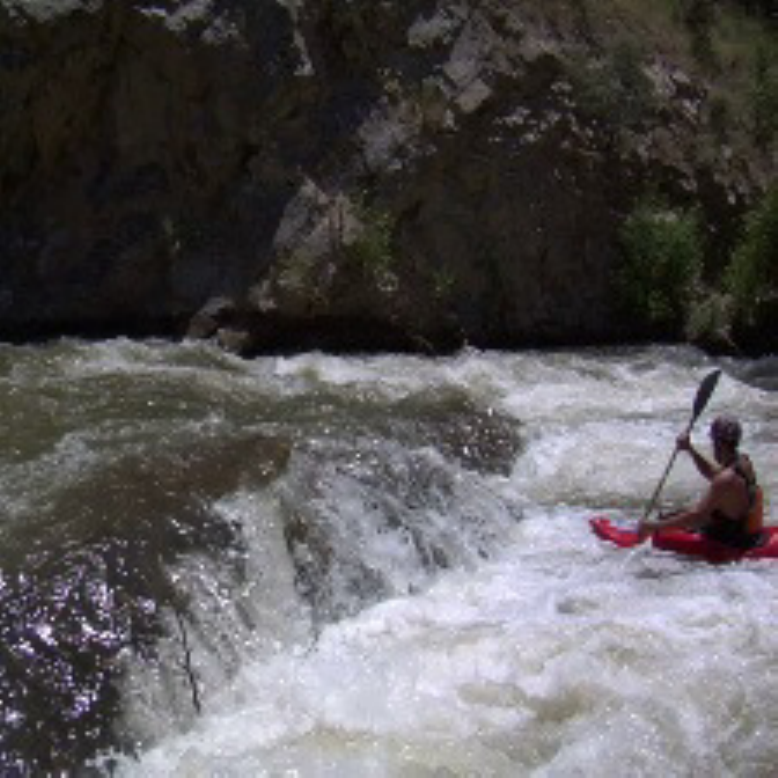}}
	\end{minipage} & \vspace{-4mm}A man is surfing in the ocean on his surf board. & \vspace{-4mm}A man is surfing in the ocean on his surf board. & \vspace{-4mm}A fire hydrant sitting on the side of an empty street. & \vspace{-4mm}A man is surfing in the ocean on his surf board. \\
	& \begin{minipage}{13mm}
	  \centering \scalebox{0.18}{\includegraphics{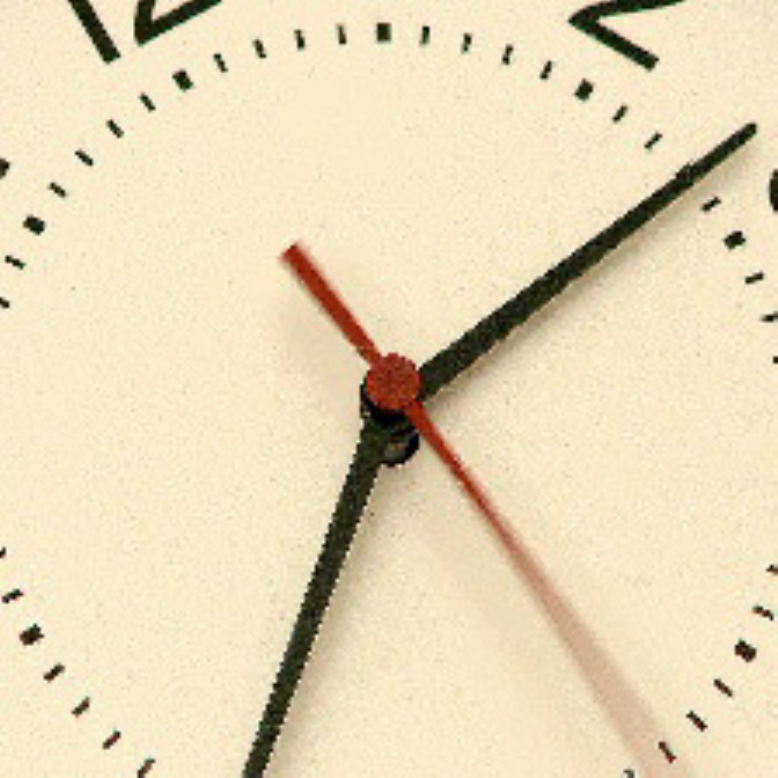}}
	\end{minipage} & \vspace{-4mm}A pair of scissors sitting on the ground. & \vspace{-4mm}A close up of an orange and white clock. & \vspace{-4mm}A fire hydrant sitting on the side of an empty street. & \vspace{-4mm}A pair of scissors sitting on the ground. \\ \hline
	\begin{minipage}{0.1mm}
	  \centering \scalebox{0.23}{\includegraphics{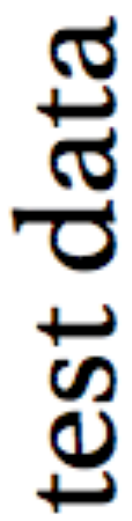}}
	\end{minipage}
	& \begin{minipage}{13mm}
	  \centering \scalebox{0.18}{\includegraphics{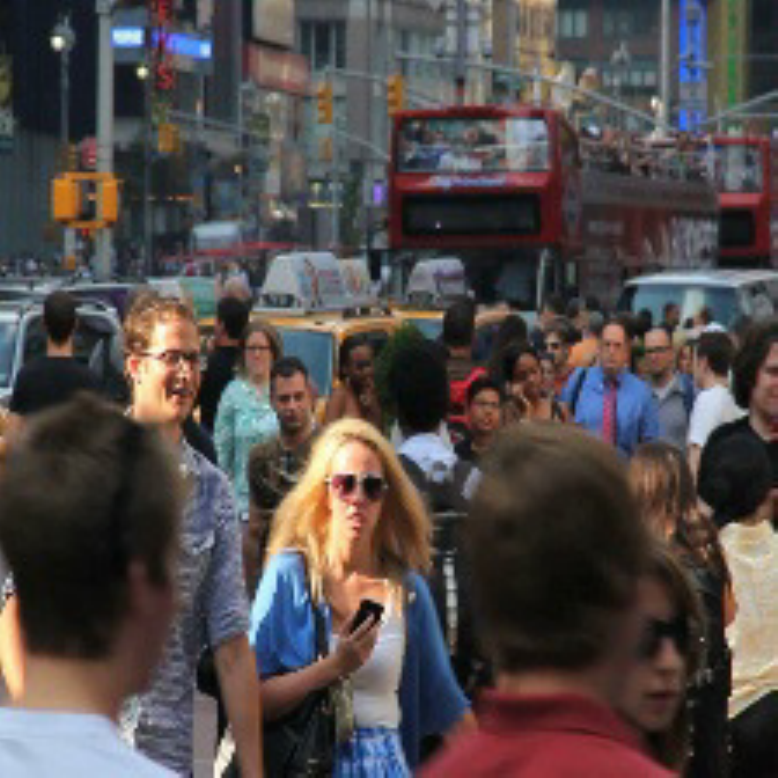}}
	\end{minipage} & \vspace{-4mm}A group of people walking down the street. & \vspace{-4mm}A group of people standing next to each other. & \vspace{-4mm}A fire hydrant sitting on the side of an empty street. & \vspace{-4mm}A group of people standing next to each other. \\
	& \begin{minipage}{13mm}
	  \centering \scalebox{0.18}{\includegraphics{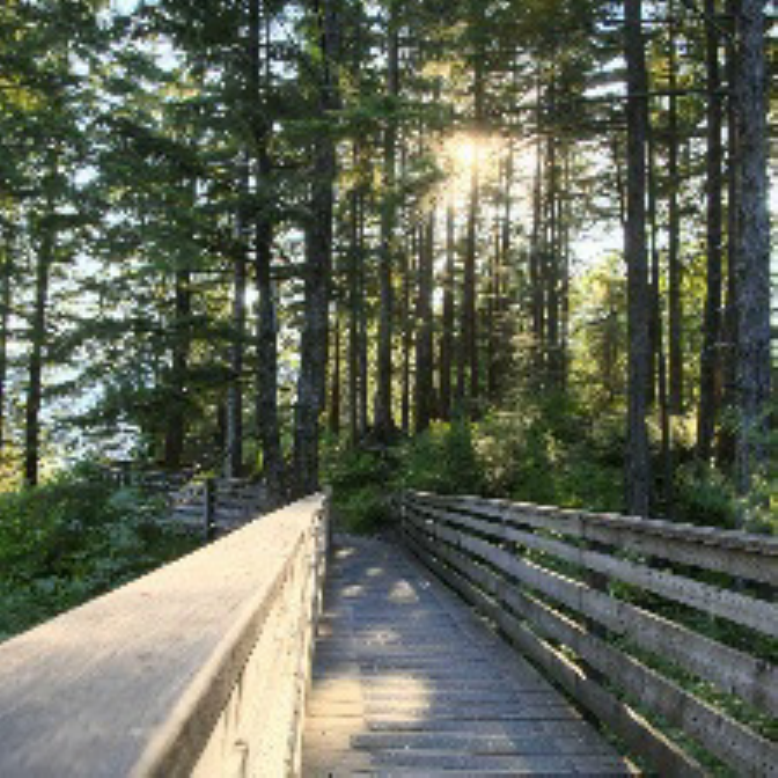}}
	\end{minipage} & \vspace{-4mm}A bench sitting in the middle of an open field. & \vspace{-4mm}A man walking down the street with an umbrella. & \vspace{-4mm}A fire hydrant sitting on the side of an empty street. & \vspace{-4mm}A train traveling down tracks next to trees. \\
       \end{tabular}
      }
    \caption{Exp. (C-1): Stimulation images and generated descriptions}
 \vspace{-1mm}
 \end{center}
\end{figure}

\subsubsection{Results \& Discussion}
The natural language descriptions generated from the four brain activity data samples (i.e., two training data and two test data), and their images are shown in Figure \ref{fig:exp_c1_brcap}. 
To compare the results, we also show the captions generated using the model in Experiment (A).
The models were evaluated with BLEU~\citep{bleu} and
METEOR~\citep{meteor}, which are the most commonly automatic evaluation
metrics in the caption generation literature (see, Figure \ref{fig:exps_c12_metrics}).
However, there are not available groundtruth sentences, i.e., human
generated descriptions for the images that evoked brain activities, which are
necessary for computing these metrics. 
Thus, we generated captions directly from 300 image samples of the test brain
dataset using the image $\rightarrow$ caption model in Experiment (A)
and then selected 60 samples among those whose generated sentences were evaluated as appropriate by a human.
For each image sample, we regarded 10 sentences generated by 10-best beam search as groundtruth data.
Note that there are several criticisms made of the BLEU and METEOR evaluation metrics, although they are considered standard metrics for translation and captioning tasks.

Human understandable natural language descriptions were generated using only brain activity data. 
As well as the model in Experiment (A), noun phrases or sentences with correct grammar, 
including prepositions and articles, were generated stably. 
The descriptions generated using the brain activity data and those using the images were nearly the same for the models employing ridge regression and the three-layer neural network. 
Thus, we confirm that learning the corresponding relationships between brain activity data and image features was successful, and the proposed method functioned well.
Taking into account of the results of Experiment (B-1), 
it was considered natural to find the same sentences were generated
for all input information in the case of using the five-layer DNN
model, and the quality of the generated sentences was low even when
employing ridge regression or the three-layer neural network to learn
the relationship.
In addition, overfitting also occurred using the five-layer DNN without pre-training with autoencoders.
Furthermore, as discussed relative to Experiment (A), the caption-generation model was learned properly, however the descriptions generated directly from the images were somewhat improper for the second test example
because the quality of the images differed, i.e., images in Microsoft COCO 
dataset were prepared for image recognition. 
Therefore, the content of the images was considerably understandable and
describable using natural language. On the other hand, the natural movies
were single-shot pictures of various types of movies that included blurring, 
darkening, letters, animation, etc. 
The content of some of these images was difficult to describe 
using natural language.
Interestingly, a proper caption was generated using brain
activity data with the three-layer neural network model compared to the image-captioning model for the second training example.
It is unlikely that a human would confuse a clock with a pair of scissors.
However, the image-captioning model made this mistake due to image processing errors in VGGNet. 
Thus, in this case, we assume that the image features obtained using brain activity data worked better than the features obtained directly from the images.

\subsection{Experiment (B-2): brain activity data $\rightarrow$ image feature model}
In Experiment (C-1), the model trained in Experiment (B-1) was overfitted 
because there was an insufficient amount of brain activity data and the dimension of the data was very large.
Therefore, we conducted an experiment 
using parts of the brain that react to visual stimuli for dimension reduction, and then trained brain activity data $\rightarrow$ image feature models (B) for higher quality.

\subsubsection{Experimental settings}
For the training dataset, 
among the 65,665 voxels of the brain activity data evoked by the visual stimuli used in Experiment (B-1),
the particular areas of the cerebral cortex used for image processing were selected as
input information. 
\citet{nishida2015} built a model that learns 
the corresponding relationships between brain activity data evoked
by visual stimuli, i.e., images, and 
the distributed semantics based on a skip-gram 
that represents the content of images. They predicted the accuracy
of how much each voxel contributes to the prediction of the corresponding relationships 
by means of the correlation co-efficiency obtained using the learned model. 
In this study, with predication accuracy thresholds of c.c. = 0.05, 0.1, 
0.15, and 0.2, useful voxels for image processing were selected to predict
semantic representations, and the number of voxels for each threshold
setting was 21,437, 9,923, 5,961, and 3,539, respectively. 
In contrast, as higher dimensional input information, we also conducted an experiment with 
the brain data whose dimensionality is 89,206 by adding the subcortex area, which 
primarily governs memory and space perception, in addition to the cerebral cortex.
Here, the three-layer neural network was used as the learning model and the experimental settings
were the same as those of Experiment (B-1).

\subsubsection{Results \& Discussion}

\begin{wrapfigure}{r}{4.0cm} 
  \vspace*{-\intextsep} 
  \includegraphics[width=4.0cm]{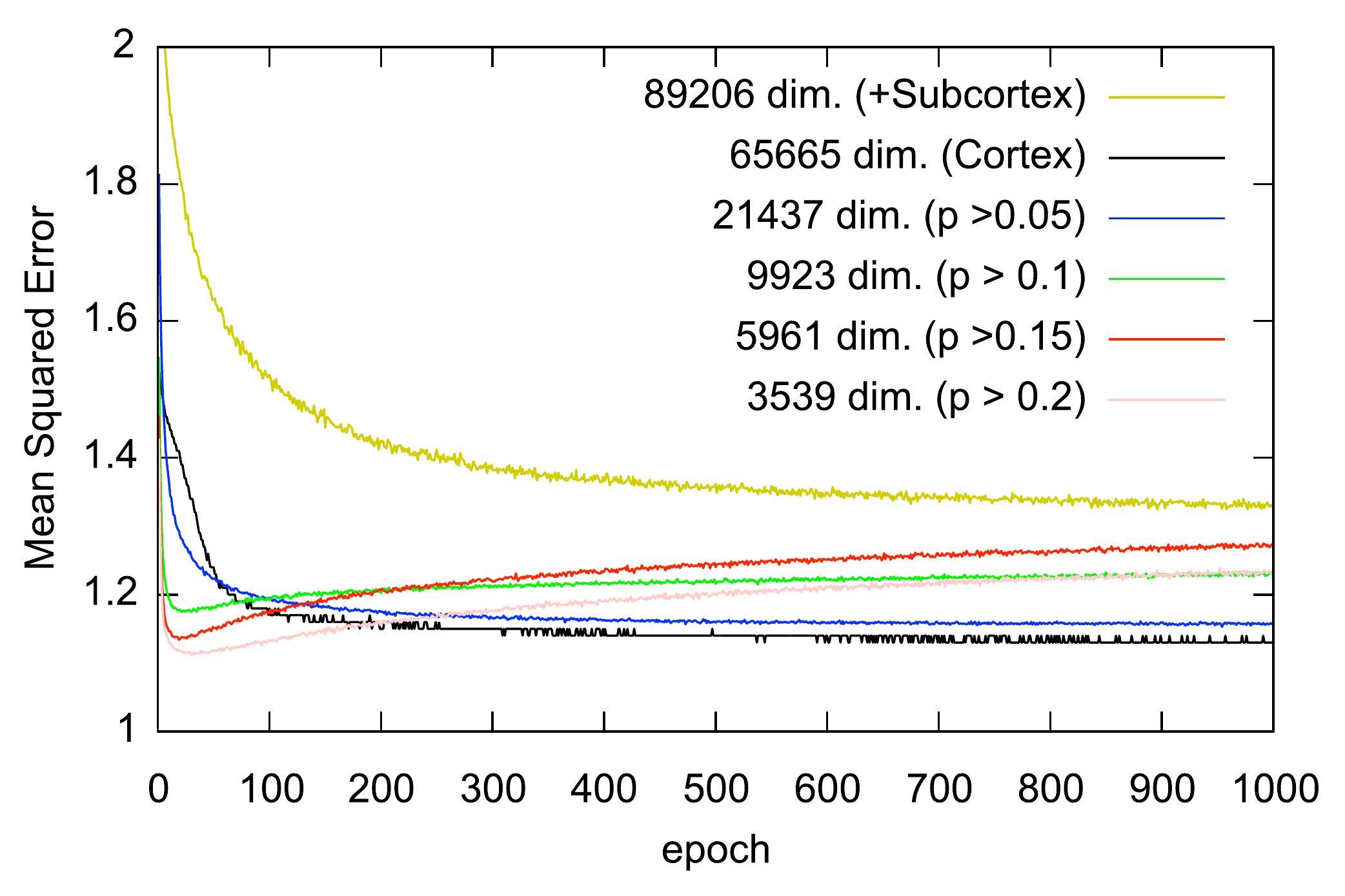}  \vspace{-7mm}
   \caption{Exp. (B-2): Changes in evaluation metrics MSE while training \label{fig:exp_b2_metrics}}
  \vspace{-5mm}
\end{wrapfigure}

Figure \ref{fig:exp_b2_metrics} shows the changes of the MSEs on the test dataset for each epoch.

In the experiments with 89,206 and 21,437 dimensional data, 
the MSEs (approximately 1.33 and 1.16, respectively) were nearly 
the same as that obtained with 65,665 dimensional data, i.e., 
Experiment (B-1) with the three-layer neural network (approximately 1.11).
On the other hand, for the experiments with 9,923, 5,961, and 21,437
dimensional data, we confirmed that overfitting occurred. The
minimum errors were 1.17 (24 epochs), 1.13 (30 epochs), and
1.10 (30 epochs), respectively. We obtained the best MSE result 
using 3,539 dimensional data. We compared and discussed six types 
of models in the following section.

\begin{figure}[t]
  \begin{center}
    \vspace{-6mm}
       \scalebox{0.63}[0.68]{
       \begin{tabular}{l p{8em} p{8em} p{8em} p{8em} p{9em} p{9em}}
	& 3,538 voxels & 5,961 voxels & 9,923 voxels & 21,437 voxels & 65,665 voxels & 89,206 voxels \\
	& ($c.c. > 0.2$) & ($c.c. > 0.15$) & ($c.c. > 0.1$) & ($c.c. > 0.05$) & (all cortex) & (+subcortex) \\ \hline
	\begin{minipage}{14.5mm}
	  \scalebox{0.22}{\includegraphics{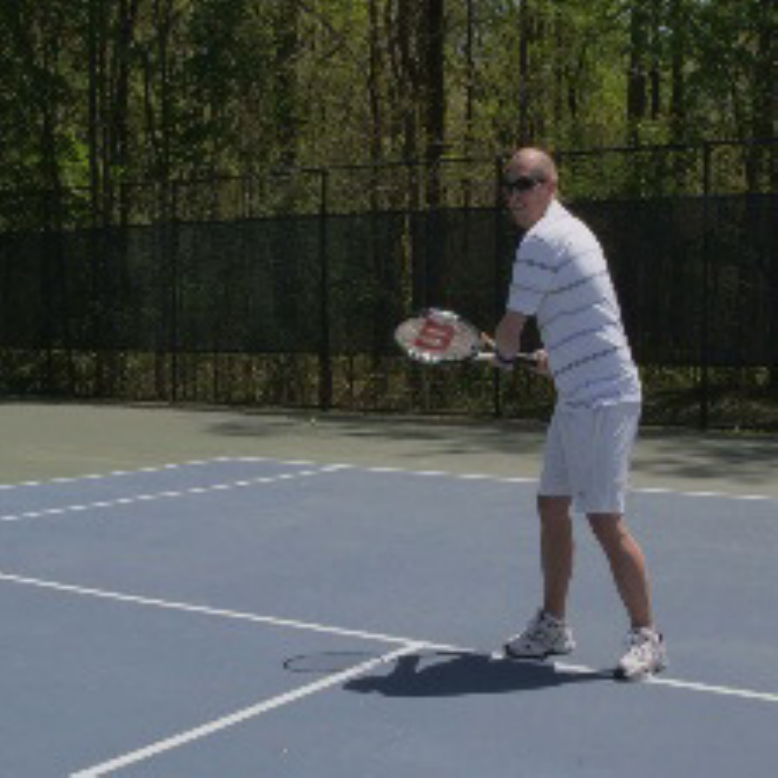}}
	\end{minipage} & \vspace{-6mm}A young man is doing tricks on his skateboard. & \vspace{-6mm}A man is playing tennis on the court. & \vspace{-6mm}A young man is playing tennis on the court. & \vspace{-6mm}A man is playing tennis on the court. & \vspace{-6mm}A man is playing tennis on the court with his racket. & \vspace{-6mm}A man is playing tennis on the court with his racket. \\
	\begin{minipage}{14.5mm}
	  \scalebox{0.22}{\includegraphics{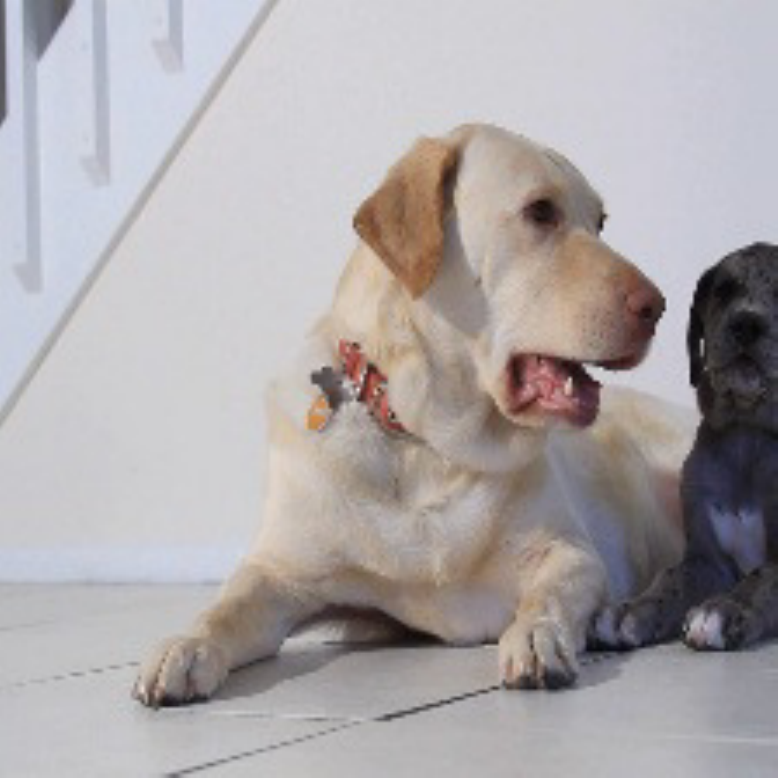}}
	\end{minipage} & \vspace{-6mm}A man sitting on the ground with an umbrella. & \vspace{-6mm}A polar bear is standing in the water. & \vspace{-6mm}A dog laying on the ground next to an orange frisbee. & \vspace{-6mm}A dog laying on the ground next to an orange frisbee. & \vspace{-6mm}A black and white dog laying on the ground. & \vspace{-6mm}A dog is sitting on the floor in front of an open door. \\
       \end{tabular}
      }
\caption{Exp. (C-2): Stimulation images and generated descriptions \label{fig:exp_c2_brcap}}
 \vspace{-6mm}
 \end{center}
\end{figure}

\subsection{Experiment (C-2): brain $\rightarrow$ caption model}
\subsubsection{Experimental settings}
Six natural language sentences were generated from brain activity data 
by combining the model in Experiment (A) and the six models in Experiment (B-2).
For the three models with 9923, 5961, and 3539 dimensions, in which overfitting occurred, 
we employed the parameters at the epoch when MSE was the lowest score.

\begin{figure}[h]
 \begin{center}  
  \includegraphics[width=10cm]{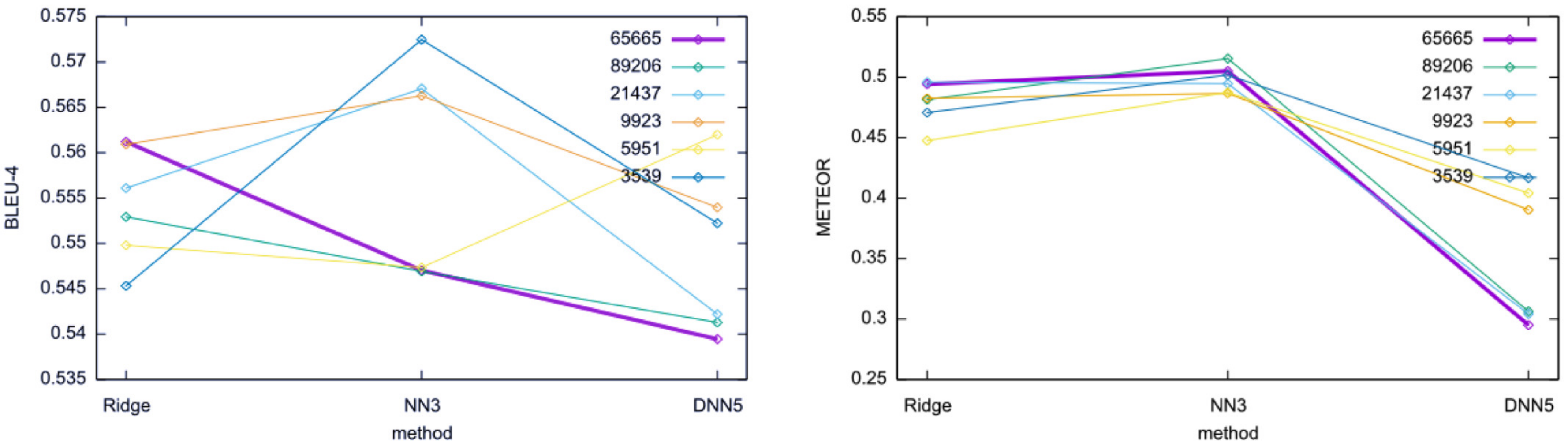}
  \vspace{-3mm}
   \caption{Exps. (C-1) \& (C-2): BLEU-4 \& METEOR scores \label{fig:exps_c12_metrics}}
  \vspace{-4mm}
 \end{center}
\end{figure}

\subsubsection{Results \& Discussion}
Six descriptions obtained for the two brain activity data samples selected from the training data.
Those descriptions and their images are shown in Figure \ref{fig:exp_c2_brcap}.
Figure \ref{fig:exps_c12_metrics} shows BLEU-4 and METEOR scores computed in the same way as Experiment (C-1).

Against expectation, it was observed that the model trained with complete
cerebral cortex data tended to produce better descriptions than the other models.
In contrast, improper descriptions were generated with the data
from the particular region of the cerebral cortex. This was also observed
when using the model obtained at 100 epochs.
The quality of the descriptions with specific brain regions was worse than that of the model trained with the whole cortex data, nevertheless learning the corresponding relationships between brain activity and visual stimuli should become easier when handling low dimensional data.
Thus, we assume that the information required to predict semantic information to understand images is included in brain areas that barely react to visual stimuli.
This result agrees with the result obtained by \citet{cukur2013}, i.e., 
most brain regions are modulated by image processing 
rather than particular regions of the cortex. 
Moreover, this also demonstrates that neural networks are effective 
for extracting important features from high dimensional data. 

As for the evaluation metrics, we confirmed that the BLUE-4 and METEOR
scores for five-layer DNNs, where overfitting clearly occurred, were the
lowest for most of the models. 
In addition, especially as for the METEOR score, the models with three-layer neural networks got slightly better results than those with ridge regression. This corresponds to the subjective evaluation based on generated descriptions mentioned above.

\section{Conclusions}
We have proposed a method to generate descriptions using brain activity data by employing a framework to generate captions for images using DNNs and by learning the corresponding relationships between brain activity data and the image features extracted using VGGNet. 
We constructed models based on three experimental settings for training
methods, and we were successful in generating natural language descriptions from brain
activity data evoked by visual stimuli.
The quality of the descriptions was higher
when using a three-layer neural network.
Moreover, by carefully examining the descriptions generated using models trained with data from different regions of the cortex, our results suggest that most brain regions are modulated by visual processing in the human brain.
In future, we plan to increase the amount of brain activity data, apply additional machine
learning methods, i.e., Bayesian optimization, whitening, etc., and revise the hyper-parameters to increase prediction accuracy.  
Furthermore, we would like to investigate proper objective methods to
evaluate the generated natural language descriptions.

\clearpage

\bibliography{matsuo_nips17}   
\bibliographystyle{acl_natbib}   

\end{document}